\documentclass[twoside,11pt]{article}

\usepackage{comment}
\usepackage{amsmath,amssymb,amsfonts}
\usepackage{bm}
\usepackage{mathtools}
\usepackage{mathrsfs}
\usepackage{cancel}
\usepackage{graphicx}
\usepackage{tikz}
\usetikzlibrary{bayesnet}
\usepackage{algorithm}
\usepackage{algpseudocode}
\usepackage{subcaption}
\usepackage{booktabs}
\usepackage[english]{babel}
\usetikzlibrary{positioning, arrows}
\usepackage{jmlr2e}

\DeclareMathOperator*{\argmax}{arg\,max}

\jmlrheading{1}{2000}{1-48}{4/00}{10/00}{Navid Ziaei, Behzad Nazari, Uri T. Eden, Alik Widge and Ali Yousefi}

\ShortHeadings{Latent Discriminative Generative Decoder Model}{Ziaei, Nazari, Eden, Widge and Yousefi}
\firstpageno{1}

\begin{document}

\title{A Bayesian Gaussian Process-Based Latent Discriminative Generative Decoder (LDGD) Model for High-Dimensional Data}

\author{\name Navid Ziaei  \email nziaei@wpi.edu \\
        \name Behzad Nazari \email nazari@iut.ac.ir \\
       \addr Department of Electrical \& Computer Engineering\\
       Isfahan University of Technology\\
       \AND
       \name Uri T. Eden \email tzvi@bu.edu \\
       \addr Department of Mathematics \& Statistics\\
       Boston University\\
       \AND
       \name Alik Widge\email awidge@umn.edu \\
       \addr Department of Psychiatry \& Behavioral Sciences\\
       University of Minnesota\\
       \AND
       \name Ali Yousefi \email ayousefi@wpi.edu \\
       \addr Department of Computer Science\\
       Worcester Polytechnic University\\
       }

\editor{}

\maketitle

\begin{abstract}
Extracting meaningful information from high-dimensional data poses a formidable modeling challenge, particularly when the data is obscured by noise or represented through different modalities. This research proposes a novel non-parametric modeling approach, leveraging the Gaussian process (GP), to characterize high-dimensional data by mapping it to a latent low-dimensional manifold. This model, named the latent discriminative generative decoder (LDGD), employs both the data and associated labels in the manifold discovery process. We derive a Bayesian solution to infer the latent variables, allowing LDGD to effectively capture inherent stochasticity in the data. We demonstrate applications of LDGD on both synthetic and benchmark datasets. Not only does LDGD infer the manifold accurately, but its accuracy in predicting data points' labels surpasses state-of-the-art approaches. In the development of LDGD, we have incorporated inducing points to reduce the computational complexity of Gaussian processes for large datasets, enabling batch training for enhanced efficient processing and scalability. Additionally, we show that LDGD can robustly infer manifold and precisely predict labels for scenarios in that data size is limited, demonstrating its capability to efficiently characterize high-dimensional data with limited samples. These collective attributes highlight the importance of developing non-parametric modeling approaches to analyze high-dimensional data.
\end{abstract}


\section{Introduction}
Dealing with high-dimensional data presents significant challenges, often resulting in analytical complications and increased computational demands. Dimensionality reduction techniques aim to transform complex, high-dimensional datasets into more manageable and insightful lower-dimensional representations~\citep{zebari2020comprehensive}. These techniques enhance data understanding and interpretation across various scientific domains, including but not limited to neuroscience, finance, and biology~\citep{koh2023dimensionality, ma2011principal, ali2017using, zhong2017forecasting}. For instance, in neural data analysis, dimensionality reduction simplifies the vast amount of data from brain imaging studies or neural recordings, aiding in the identification of patterns of activity underlying cognitive processes or disorders~\citep{cunningham2014dimensionality, wang2014generalized, heller2022targeted}.

Dimensionality reduction techniques fall into two main categories: parametric and non-parametric. Non-probabilistic methods include principal component analysis (PCA)~\citep{bro2014principal}, linear discriminant analysis (LDA)~\citep{balakrishnama1998linear}, and more recent approaches including t-distributed stochastic neighbor embedding (t-SNE)~\citep{cieslak2020t} and uniform manifold approximation and projection (UMAP)~\citep{mcinnes2018umap}. These methods are widely used for data simplification and visualization. Despite their significance in advancing science and their popularity, they often lack the ability to provide a sophisticated understanding, such as stochasticity in the inferred projections, and are highly sensitive to noise and outlier data points. On the other hand, probabilistic models, including probabilistic PCA (PPCA)~\citep{tipping1999probabilistic}, variational auto-encoders (VAE)~\citep{kingma2019introduction}, and Gaussian process latent variable model (GPLVM)~\citep{lawrence2005probabilistic} are gaining prominence as dominant approaches in data analysis. The GPLVM and PPCA address major shortcomings of non-probabilistic methods, as their inference of low-dimension representation is probabilistic rather than deterministic. GPLVM is built upon PPCA, with the added advantage of utilizing non-linear mappings to find lower-dimensional representations. Despite significant progress in these models, challenges persist, and there are opportunities to enhance their ability to characterize complex and high-dimensional data. A key area for improvement is the integration of label and category information, which these models often overlook due to their predominantly unsupervised nature. An important challenge facing Gaussian process-based models like GPLVM is their excessive computational cost, especially as the data size increases. Our research seeks to tackle these computational challenges and employ label and category information to enhance the scalability and utility of models like GPLVM.

To address the high computational cost associated with non-parametric models, several approaches have been developed. The Nyström approximation~\citep{williams2000using} offered an initial method to reduce computational requirements by sampling a subset of the data to approximate the Gaussian process covariance matrix. This was followed by the introduction of the pseudo-inputs concept by~\cite{snelson2005sparse}, which parameterizes its covariance using the locations of $M$ pseudo-input points, determined optimally through gradient-based optimization. Building upon these foundations,~\cite{titsias2009variational} introduced variational inducing points, utilizing variational inference for the optimal selection of inducing points. Furthermore, stochastic gradient descent-based variational inducing points, developed by~\cite{hensman2013gaussian}, employed stochastic optimization to further enhance scalability for large datasets. These advancements lay the foundation for further enhancements in this category of models.

Alongside advancements in enhancing the scalability of GPLVMs, there is growing research aimed at integrating labels or categories into the dimensionality reduction process. This integration allows for the inference of both the underlying structure and its relevance within the mapping. Incorporating the labels of data points into the dimensionality reduction process has been explored in approaches such as supervised PPCA (SPPCA)~\citep{yu2006supervised}. This approach employs two linear mappings: one mapping from the latent space to the input space, representing the original high-dimensional data, and another mapping from the latent space to the output space, corresponding to the data points' label or category or other supervised information. This concept was extended to nonlinear dimensionality reduction techniques, especially with the GPLVM approach. 
The discriminative GPLVM (D-GPLVM) applies principles of generalized discriminant analysis (GDA)~\citep{baudat2000generalized} to modify latent positions within the data, aiming to bring data points of the same class closer and distance those of differing classes. Here, latent positions denote the model's estimated, unobserved variables that capture the underlying structure of the data. Other approaches, such as the supervised GPLVM and shared GPLVM (SGPLVM)~\citep{gao2010supervised, ek2009shared, jiang2012supervised}, employ two distinct GPLVMs to uncover nonlinear relationships among latent variables, data, and labels. The integration of SGPLVM with SPPCA, also known as supervised latent linear GPLVM (SLLGPLVM), was presented by~\cite{jiang2012supervised}. This approach employs two mapping functions to link the input and output spaces via a latent variable space. To minimize computational costs, SLLGPLVM employs a linear mapping from latent variables to high-dimensional data, similar to the method used by~\cite{zhang2017supervised}. While this modification reduces the computational cost, it sacrifices the flexibility to capture complex nonlinear relationships inherent in the data, potentially limiting the model's ability to properly model highly intricate patterns found in more complex datasets. A further extension to SGPLVM is the supervised GPLVM based on a Gaussian mixture model (SGPLVM-GMM)~\citep{zhang2017supervised}. SGPLVM-GMM assumes that latent variables follow a Gaussian mixture distribution, where mixture components are conditioned on class labels. Thus, the model simultaneously learns to reduce data dimension and classify data samples by predicting class probabilities. These models, including SGPLVM and its extensions such as SLLGPLVM and SGPLVM-GMM, have successfully utilized latent variables and class labels to enhance feature extraction and data classification. However, they primarily rely on point estimates of the latent process, thereby increasing the risk of overfitting, especially in scenarios where linear relationships do not fully capture data structure. 

Although these models have made progress in incorporating training labels into latent process inferencing, a gap remains in their ability to manage the complexities introduced by noise and limited sample sizes. To address these challenges, our research aims to develop a fully Bayesian dimensionality reduction solution that integrates the samples' labels. Another challenge with previous GP-based models is scalability, which we address in this work through variational inducing points. The distinctiveness of our model lies in employing a Bayesian approach for both training and inference while assuming a prior over latent space. The Bayesian approach helps us better manage data complexity introduced by noise and missing points; it also lets the framework be applied to datasets that have a limited sample size. Our approach, which can be considered an extension of SGPLVM, presents a robust and scalable method for dimensionality reduction as we incorporate the inducing points. Through multiple examples, we demonstrate that our proposed method enhances the feature extraction process by mapping data points to a lower dimension and ensures the retention of essential data patterns. We illustrate that our proposed approach can be used as a  robust decoder model to draw low-dimensional representations and labels for new data points. In deriving the model, we use Gaussian processes (GPs) and Bayesian inference, turning the model into a non-parametric approach. As such, the solution prevents overfitting and is best suited for datasets with a limited number of samples. Our proposed approach can be readily applied to datasets from various fields. As such, we focus on highlighting the framework application and evaluating its performance through synthetic and benchmark datasets.

In this research, we introduce our method which is called  Latent Discriminative Generative Decoder (or, briefly LDGD) model. In LDGD, a low-dimensional random variable defines the underlying generative process responsible for generating both the continuous values and label data. This random variable defines the covariance structures of two Gaussian processes (GPs)---one for the continuous measurement and another for the label. We then employ a Bayesian inference solution integrated with the variational inducing points approach to train the model. The usage of inducing points enables efficient management of computational costs during training and inference. Specifically, we draw two sets of inducing points, one set for each GP. With two sets of inducing points, we show that LDGC achieves a balanced  prediction accuracy and data generation capability. Additionally, we will introduce a batch training pipeline for LDGD, which significantly aids in faster training and enhances the scalability of the model for larger datasets. LDGD also serves as an adaptive feature extraction and classification solution, allowing for the prediction of data point labels from continuous values. Moreover, it functions as a generative model capable of producing data in high-dimensional space. To facilitate further research and development, the code for LDGD implementation is openly available at our GitHub repository: \url{https://github.com/Navid-Ziaei/LDGD}. While some of these attributes are shared with variational auto-encoders, we argue that this framework represents a substantial advancement over variational auto-encoders (VAE).

A key enhancement of LDGD over VAEs is its fully Bayesian inference process, which quantifies prediction uncertainty as the posterior's variance. Another notable advancement is LDGD's ability to partially optimize the dimensions of latent variables during the training phase, enhancing both feature extraction and model interpretability. Additionally, LDGD's Bayesian framework inherently protects against overfitting, making it well-suited for complex datasets with limited sample sizes. These advantages collectively position LDGD not just as a mere alternative but as a significant advancement over traditional encoder-decoder models, especially in high-dimensional data analysis. 

The subsequent sections will explore the technical and applied aspects of the LDGD modeling framework. We first delve into the intricacies of the LDGD framework, covering its formulation, model training, and inference procedures in detail. Additionally, we introduce a variant of LDGD, termed "fast LDGD", which integrates a neural network into the LDGD framework, enhancing its computational efficiency and predictive power. To assess the performance and attributes of our proposed model, we have conducted evaluations of LDGD on a synthetic dataset. Following this, we demonstrate its applicability across several benchmark datasets, including Oil Flow, Iris, and MNIST. Subsequently, we compare the distinctive features of LDGD via other methods, such as SLL-GPLVM and SGPLVM, to highlight its unique advantages and improvements. The final part of our analysis is a comparison between LDGD and variational auto-encoders. This thorough review highlights the significant contributions of the LDGD model to machine learning, especially in efficiently reducing dimensions and accurately decoding information.

\section{Materials and Methods}
\subsection{Gaussian process Regression}
\label{section:gpr}
Gaussian processes (GP) offer a powerful non-parametric approach for regression and classification tasks  \citep{seeger2004gaussian}. A Gaussian process is a collection of random variables, with the property that any finite subset of these variables has a joint Gaussian distribution. GP defines a distribution over a function defined by: 
\begin{equation*}
p(\mathbf{f}\mid\mathbf{X}) = \mathcal{N}(\boldsymbol{\mu}, \mathbf{K}_{NN}),
\end{equation*}
where $\mathbf{X} = [\mathbf{x}_1, \ldots, \mathbf{x}_N]$ are input vectors with $\mathbf{x}_i \in \mathbb{R}^d $, $\mathbf{f} = [f(\mathbf{x}_1), \ldots, f(\mathbf{x}_N)]$ represents latent variables, called function values in GP context, at the input points with mean $\boldsymbol{\mu} = [m(\mathbf{x}_1), \ldots, m(\mathbf{x}_N)]$ and a $N\times N$ covariance matrix $\mathbf{K}_{NN}$. 
Each element in the $i$th row and $j$th column is given by $K_{ij} = k_\theta (\mathbf{x}_i, \mathbf{x}_j)$ for $i, j \in \{1, \ldots, N\}$, where $k$ is a positive definite kernel constructing elements of the covariance matrix $\mathbf{K}_{NN}$ and $\theta$ defines the kernel's free parameters. In practice, with no prior observation, the mean of the function is assumed to be zero, i.e., $m(\mathbf{x}) = \mathbf{0}$.

Consider a dataset $d = \{(\mathbf{x}_i, y_i)\}_{i=1}^{N}$, where $y_i \in \mathbb{R}$ represents the noisy observations at each input location $\mathbf{x}_i$ within the domain of interest, encapsulating the underlying process we aim to model using a Gaussian process. The generation of $y_i$ is modeled as follows:
\begin{equation*}
y_i = f(\mathbf{x}_i) + \epsilon,
\end{equation*}
where $\epsilon$ is considered as Gaussian noise, defined by $\epsilon \sim \mathcal{N}(0, \sigma^2_y)$. The likelihood of observing $y_i$ given $\mathbf{x}_i$ is defined by
\begin{equation}
p(y_i \mid \mathbf{x}_i) = \mathcal{N}(y_i \mid f(\mathbf{x}_i), \sigma^{2}_{y}).  
\end{equation} 
Furthermore, given the GP model, the joint distribution of $\mathbf{y}=[y_1, \dots, y_N]$ given $\mathbf{X}$ is defined by
\begin{equation*} 
p(\mathbf{y}\mid \mathbf{X}) = \mathcal{N}(\mathbf{y}\mid\mathbf{0}, \mathbf{K}_{NN} + \sigma^2_y \mathbb{I}_N) .  
\end{equation*} 
To train the model, we need to estimate the kernel parameters ($\theta$), which can be found by solving
\begin{equation*} 
\arg\max_\theta \log p(\mathbf{y}\mid \mathbf{X}).  
\end{equation*} 
For a set of test points $\mathbf{X}^*\equiv [\mathbf{x}^*_1, \ldots, \mathbf{x}^{*}_{N^*}]$, we are interested in finding the predictive distribution $p(\mathbf{y}^*\mid\mathbf{y}, \mathbf{X}, \mathbf{X}^*) = \mathcal{N}(\boldsymbol{\mu}^*, \boldsymbol{\Sigma}^*)$, where the mean and covariance are defined by
\begin{align*}
    \boldsymbol{\mu}^* &= \mathbf{K}_{*N} \mathbf{K}_{NN}^{-1}\mathbf{y},  \\
    \boldsymbol{\Sigma}^* &= \mathbf{K}_{**} - \mathbf{K}_{*N}(\mathbf{K}_{NN}+\sigma^{2}_{y}\mathbb{I}_N)^{-1} \mathbf{K}_{N*} .
\end{align*} 
Here, $\mathbf{K}_{**}$ is the covariance matrix between the test points, and $\mathbf{K}_{*N}$ is the cross-covariance matrix between the test points and the training points. 

The formulation requires the inversion of an $N \times N$ matrix to derive the predictive distribution, resulting in a computational complexity of $\mathcal{O}(N^3)$. This computation becomes expensive in datasets even with a moderate number of data points. A solution that addresses this challenge is based on variational inducing points proposed by \cite{titsias2009variational}. In this approach, $M$ data points $\mathbf{Z} \equiv \{\mathbf{z}_i\}_{i=1}^{M}$, where $\mathbf{z}_i \in \mathbb{R}^d$, is introduced in the input space. These $M$ data points, where $M \ll N$, replace the $N$ original data points used in deriving the covariance matrix. Inducing points $\mathbf{Z}$ can be chosen as fixed points or model-free parameters \citep{titsias2009variational}. This research considers $\mathbf{Z}$ as free parameters with random initial values. We further assume for each inducing point, there is a corresponding inducing variable $u_i$. Induced variables $\mathbf{u} \equiv \left[u_1, ..., u_M\right]$ follow the distribution $p(\mathbf{u}\mid\mathbf{Z}) = \mathcal{N}(\mathbf{u} \mid \mathbf{0}, \mathbf{K}_{MM})$, where $\mathbf{K}_{MM}$ is an $M \times M$ matrix. This matrix is generated using the same kernel function $k_{\theta}(z_i, z_j)$ as in the GP model, applied to the inducing points. The conditional distribution of $\mathbf{f}$ given $\mathbf{u}$ is:
\begin{equation}
\label{eq_spgr:predictivedist}
    p(\mathbf{f} \mid \mathbf{u}; \mathbf{Z}, \mathbf{X}) = 
    \mathcal{N}(\mathbf{f} \mid \boldsymbol{\mu_f}, \boldsymbol{\Sigma_f}), 
\end{equation} 
where $\boldsymbol{\mu_f} = \mathbf{K}_{NM}\mathbf{K}_{MM}^{-1}\mathbf{u}$ and $\boldsymbol{\Sigma_f} = \mathbf{K}_{NN} - \mathbf{K}_{NM} \mathbf{K}_{MM}^{-1} \mathbf{K}_{MN}$.  
Using the inducing points, we can draw predictions for new points with a much smaller dataset and significantly reduce the computational cost for the covariance inverse. However, to use $\mathbf{Z}$ for the prediction step, we first need to estimate these points. With the inducing points, the free parameters of the GP model increase to include both $\theta$ and $\mathbf{Z}$. We find the maximum likelihood estimate of $\theta$ and $\mathbf{Z}$ through the marginal distribution of $\mathbf{y}$, defined by
\begin{align*}
    p(\mathbf{y} \mid \mathbf{X} ;\mathbf{Z}, \theta) 
    &= \int 
        p(\mathbf{y}, \mathbf{f}, \mathbf{u} \mid \mathbf{X}; \mathbf{Z}, \theta) 
        \, d\mathbf{f} \, d\mathbf{u} \notag \\
    &= \int 
        p(\mathbf{y} \mid \mathbf{f}; \theta) 
        p(\mathbf{f} \mid \mathbf{u}, \mathbf{X}; \mathbf{Z}, \theta) 
        p(\mathbf{u}; \mathbf{Z}, \theta) \, d\mathbf{f} \, d\mathbf{u},
\end{align*}
where $\mathbf{f}$ and $\mathbf{u}$ are set of latent variables of length $N$ and $M$, respectively. To determine $\theta$ and $\mathbf{Z}$ without needing to invert the large matrix $\mathbf{K}_{NN}$, we employ a variational approach by introducing the variational distribution $q(\mathbf{f}, \mathbf{u})$. We define $q(\mathbf{f}, \mathbf{u})$ as $q(\mathbf{u})q(\mathbf{f} \mid \mathbf{u})$, where $q(\mathbf{u}) = \prod_{j=1}^{M} \mathcal{N}(u_j; m_j, s_j)$ and $q(\mathbf{f} \mid \mathbf{u}) = p(\mathbf{f} \mid \mathbf{u}, \mathbf{X}; \mathbf{Z}, \theta)$, as already defined in Equation~\eqref{eq_spgr:predictivedist}. With the $q$ distribution, we can derive a lower bound on the marginal distribution of $\mathbf{y}$ using Jensen inequality, which becomes:
\begin{align}
    \log p(\mathbf{y} \mid \mathbf{X}; \mathbf{Z}, \theta) 
    &\geq \mathbb{E}_{q(\mathbf{f}, \mathbf{u})}  
    [ \log p(\mathbf{y} \mid \mathbf{f}; \theta )] 
    - \text{KL}[q(\mathbf{u}) \parallel p(\mathbf{u}; \mathbf{Z}, \theta)],
    \label{eq:sgpr_elbo}
\end{align}
where we find $\mathbf{Z}$ and $\theta$, along with $m_j, s_j$ for $j = 1, \ldots, M$ to maximize the lower bound. The first term on the right-hand side of the Equation \ref{eq:sgpr_elbo} encourages the model to fit the data well, while the second term regularizes the model by penalizing deviations of the variational distribution from the prior. It is worth noting that the inversion of the $\mathbf{K}_{NN}$ matrix is no longer needed in prediction and parameter estimation steps.

For $N_t$ test points $\mathbf{X}^* \equiv \left[ x^*_1, ..., x^*_{N_t} \right]$, the predictive distribution is defined by $p(\mathbf{f}^* \mid \mathbf{X},\mathbf{y},\mathbf{X}^*)$. Given the inducing points, we can write this distribution as:
\begin{equation*}
    p(\mathbf{f}^* \mid \mathbf{X},\mathbf{y},\mathbf{X}^*) = 
    \int p(\mathbf{f}^*, \mathbf{u} \mid \mathbf{X},\mathbf{y},\mathbf{X}^*) \, 
    d\mathbf{u}.
\end{equation*}
Note that in the equation, for clarity, we dropped its parameters - i.e., $\mathbf{Z}$ and $\theta$. We estimate the exact posterior $p(f^*,\mathbf{u} \mid \mathbf{X},\mathbf{y},\mathbf{X}^*)$ with the updated variational distribution $q(f^*,\mathbf{u})$. Thus, the predictive distribution can be approximated as:
\begin{align}
    p(\mathbf{f}^* \mid \mathbf{X},\mathbf{y},\mathbf{X}^*) 
    &\approx \int q(\mathbf{f}^*,\mathbf{u}) \, d\mathbf{u} = \int p(\mathbf{f}^* \mid \mathbf{u}) q(\mathbf{u}) \, d\mathbf{u}. 
    \label{eq:integral_approx}
\end{align}

We have already derived these distributions during the model training. They are defined by:
\begin{align*}
    q(\mathbf{u}) 
    &\sim \mathcal{N}(\mathbf{u} \mid \mathbf{m}, \mathbf{S}), \\
    p(\mathbf{f}^* \mid \mathbf{u}) 
    &\sim \mathcal{N}
    \left(
        \mathbf{f}^* \mid \mathbf{K}_{*M} \mathbf{K}_{MM}^{-1} \mathbf{u}, \mathbf{K}_{**} - 
        \mathbf{K}_{*M} \mathbf{K}_{MM}^{-1} \mathbf{K}_{M*}
    \right).
\end{align*}
where $\mathbf{m}=[m_1, \ldots ,m_M]$ and $\mathbf{S}$ is a $M \times M$ diagonal matrix. The diagonal elements of $\mathbf{S}$ are $\{ s_i\}_{i=1}^{M}$, which are the variational free parameters that are optimized during training. The approximated integral in Equation~\eqref{eq:integral_approx} has a closed form, which is a Gaussian with the following mean and covariance:
\begin{align*}
    \boldsymbol{\mu}^* &= \mathbf{K}_{*M} \mathbf{K}_{MM}^{-1} \mathbf{m}, \\
    \boldsymbol{\Sigma}^* &= \mathbf{K}_{**} - \mathbf{K}_{*M} \mathbf{K}_{MM}^{-1} (\mathbf{K}_{MM} - S) \mathbf{K}_{MM}^{-1} \mathbf{K}_{M*}.
\end{align*}

\subsection{Gaussian Process Latent Variable Model}
Gaussian process latent variable models emerge from the necessity of understanding high-dimensional data by projecting it into a simpler and much lower-dimensional latent space. At its core, a GPLVM is built upon the hypothesis that even though we observe data $\mathbf{Y} \in \mathbb{R}^{N \times D}$, there exists a latent representation $\mathbf{X} \in \mathbb{R}^{N \times Q}$, with $Q\ll D$, that captures the essential features or structures present in the high-dimensional data. This idea was inspired by PPCA, evolving into a nonlinear dimensionality reduction method that extends the capabilities of its precursor \citep{tipping1999probabilistic}. The relationship between GPLVM and PPCA is explained in Appendix~\ref{appendix_ppca}. In GPLVM, the conditional distribution of $\mathbf{Y}$ given $\mathbf{X}$ is assumed to be:
\begin{equation*}
    p(\mathbf{Y} \mid \mathbf{X}) = \prod_{d=1}^{D} p(\mathbf{y}_{:,d} \mid \mathbf{X})\text{,}
\end{equation*}
where $\mathbf{y}_{:,d}$ is $d $th column of $\mathbf{Y}$. Each element of $\mathbf{y}_{:,d}$ is a noisy realization of the function $\mathbf{f}_{d} \in \mathbb{R}^N$ at location $\mathbf{x}_i$ defined by 
\begin{equation*}
    y_{i,d} = f_d(\mathbf{x}_i) + \epsilon_d,
\end{equation*}
where $\epsilon_d \sim \mathcal{N}(0,\sigma^2_d)$ and $\mathbf{f}_{d}\equiv \left[f_d(\mathbf{x}_1), ..., f_d(\mathbf{x}_N) \right]$ is function values with GP prior. This leads to the following conditional distribution:
\begin{equation*}
    p(\mathbf{y}_{:,d}\mid \mathbf{X}) = \mathcal{N}(\mathbf{y}_{:,d} \mid \mathbf{0}, \mathbf{K}_{NN} + \sigma_d^2 \mathbb{I}_N),
\end{equation*}
with $\mathbf{K}_{NN}$ being an $N \times N$ covariance matrix defined using a kernel function as a function of the latent space $\mathbf{X}$. 

We use the ARD kernel \citep{seeger2004gaussian} for the kernel. This kernel is generally used for determining the relevance of dimensions in $\mathbf{X}$ by tuning ${\alpha}_q$. A typical ARD kernel is defined by:
\begin{equation*}
k(\mathbf{\mathbf{x}}, \mathbf{\mathbf{x}'}) = \sigma_f^2 \exp\left(-\frac{1}{2} \sum_{q=1}^{Q} \alpha_q (x_q - x'_q)^2\right),
\end{equation*}
where $\mathbf{x}$ and $\mathbf{x'}$ represent two input vectors in the latent space. The term $\sigma_f^2$ denotes the variance parameter of the kernel, dictating the scale of the output space. Each $\alpha_q$ is an inverse length-scale parameter associated with the $q$th dimension of the latent space, which influences how variations in that particular dimension affect the kernel computation. These length-scale parameters allow the kernel to adjust the relevance of the feature dimensions.

The GPLVM can be interpreted as a variant of a multi-output Gaussian process. Every dimension of the observed data, $\mathbf{Y}$, is a separate realization of a Gaussian process, all derived by a common latent structure \citep{liu2018remarks}. To fit GPLVM to a dataset, we require finding a suitable representation of the latent variables that best explain the observed data. This can be done by estimating ARD kernel-free parameters. Three primary methods used for finding suitable representation include point estimate, Maximum A Posteriori (MAP) \citep{lawrence2005probabilistic}, and a fully Bayesian approach \citep{titsias2010bayesian}. Whereas the point estimate method can be more susceptible to overfitting, the Bayesian approach typically offers a better generalization to data due to its integration over uncertainties. The payoff is that the Bayesian approach is computationally more expensive and often requires sophisticated inference techniques such as Markov Chain Monte Carlo (MCMC) \citep{gilks1995markov} or variational inferencing \citep{blei2017variational}. This research focuses on Bayesian methods to derive a more robust inference. In the Bayesian framework, it is imperative to specify a prior distribution for the latent variables. We assume the prior on the latent variables are Gaussian defined by:
\begin{equation*}
    p(\mathbf{X}) = \prod_{n=1}^{N} \mathcal{N}(\mathbf{x}_n \mid \mathbf{0}, \mathbb{I}_Q),
\end{equation*}
where $\mathbf{x}_n$ denotes the $n$th row of the latent variables matrix $\mathbf{X}$. With this prior, the marginal probability of $\mathbf{Y}$ involves the integral, which is defined by:
\begin{equation}
     p(\mathbf{Y}) 
     = \int{p(\mathbf{Y}\mid\mathbf{X})p(\mathbf{X})d\mathbf{X}} = \int{\prod_{d=1}^{D}p(\mathbf{y}_{:,d}\mid \mathbf{X}) 
     p(\mathbf{X})d\mathbf{X}}.
\label{eq:sgpr_evidence}
\end{equation}
This integral corresponds to the expectation of conditional probability of $\mathbf{Y}$ over $\mathbf{X}$. The integral does not have a closed-form solution, as $\mathbf{K}_{NN}$ is a function of $\mathbf{X}$. To find the marginal distribution, we apply Jensen's inequality, which gives a variational lower bound on the log of likelihood:
\begin{equation*}
    \log{p(\mathbf{Y})} \geq \sum_{d=1}^{D}{\mathbb{E}_{q(\mathbf{X})}[\log p(\mathbf{y}_{:,d}\mid\mathbf{X})]} - \text{KL}(q(\mathbf{X}) \parallel p(\mathbf{X}\mid\mathbf{Y})).
\end{equation*}
It is necessary to pick a proper $\mathbf{q}$ distribution to achieve a tight bound here. The number of samples in the training set introduces additional complexities to the model; thus, we will use the inducing points to derive a more tractable solution to calculate the bound. Different strategies have been proposed to calculate the bound, including the solution proposed by \cite{hensman2013gaussian}, which creates a closed-form solution for the bound. \cite{salimbeni2017doubly} employed the variational distribution to simplify the problem by enabling lower-dimensional sampling conducive to convergence. Here, we discuss our approach, which integrates the labels or categories attached to each data point into the formation of our low-dimensional representation, thereby facilitating the classification of these labels. This has been accomplished through the doubly variational method, upon which we have developed the LDGD.

\subsection{Latent Discriminative Generative Decoder Model}
Here, we introduce the LDGD framework. While the idea behind LDGD is the same as GPVLM, our objective is different. In LDGD, we aim to infer the latent process that can capture essential features of the observed data and separate the classes in the data, contrasting with GPLVM, which primarily focuses on dimensionality reduction without explicitly optimizing for class separability. Essentially, we are identifying a set of features in the data that can be used for both data generation and achieving high accuracy in the data class or category prediction. We note that LDGD inference will differ from those derived by generative auto-encoders, as the inferred latent process needs to capture those essential features of the data that are expressive of labels or categories of data. The following section will delve into the framework definition and its different modeling steps, including training and prediction.

\subsubsection{Observed Data}
In LDGD, we assume the observed data consists of continuous measures, represented as $\mathbf{Y}^r \in \mathbb{R}^{N \times D}$, and corresponding classes of data or categories, represented as $\mathbf{Y}^c \in \mathbb{R}^{N \times K}$. $N$ denotes the number of samples, $D$ is the dimension of continuous measures per sample, and $K$ is the number of distinct categories or classes per sample. Each row of $\mathbf{Y}^r $, denoted as $\mathbf{y}^r_i$, corresponds to the continuous measurement for the $i$th sample. Each column, denoted as $\mathbf{y}^r_{:,d}$, represents the data for the $d $th dimension across all samples. Similarly, each row of $\mathbf{Y}^c$, denoted as $\mathbf{y}^c_{i}$, corresponds to the one-hot encoded class label for each sample.

\subsubsection{Latent Variables}
LDGD incorporates latent variables $\mathbf{X} \in \mathbb{R}^{N \times Q}$, to capture the essential features driving both $\mathbf{Y}^r $ and $\mathbf{Y}^c$. Here, $Q$ represents the latent space's dimensionality, which is generally much smaller than the dimension of the observed data. We assume a Gaussian prior for these latent variables, expressed as:

\begin{equation}
    p(\mathbf{X}) = \prod_{i=1}^{N} \mathcal{N}(\mathbf{x}_i \mid \mathbf{0}, \mathbb{I}_Q),
    \label{eq:BGPLVM_prior}
\end{equation}
where $\mathbf{0}$ is a zero vector and $\mathbb{I}_Q$ is the identity matrix of size $Q\times Q$. 

\subsubsection{Gaussian process Priors}
To model the relationships between the latent space and the observed data, we impose two Gaussian process (GP) priors on function values $\mathbf{F}^r $ and $\mathbf{F}^c$. Specifically, we define GP priors over $\mathbf{F}^r \in \mathbb{R}^{N \times D}$, for the continuous measurements, and $\mathbf{F}^c \in \mathbb{R}^{N \times K}$ for the categorical measurements or labels. These prior distributions are defined by:
\begin{align}
    p(\mathbf{F}^c \mid \mathbf{X}) &= \prod_{k=1}^{K} \mathcal{N}(\mathbf{f}_{:,k}^c \mid \mathbf{0}, \mathbf{K}_{NN}^c) \label{eq:prior1} \\
    p(\mathbf{F}^r \mid \mathbf{X}) &= \prod_{d=1}^{D} \mathcal{N}(\mathbf{f}_{:,d}^r \mid \mathbf{0}, \mathbf{K}_{NN}^r),
    \label{eq:prior2}
\end{align}
where $\mathbf{K}_{NN}^c$ and $\mathbf{K}_{NN}^r $ are the $N\times N$ covariance matrices for the GPs corresponding to class labels and observed measurements, respectively. The element in the $i$th row and $j$th column of these matrices are defined by kernel functions $k^{c}_{i,j}=k_{\psi}(\mathbf{x}_i,\mathbf{x}_j)$ and $k^{r}_{i,j}=k_{\theta}(\mathbf{x}_i,\mathbf{x}_j)$, with $\theta$ and $\psi$ as the kernel parameters. Furthermore, $\mathbf{f}_{:,k}^c$ represents the $k$th column of $\mathbf{F}^c$, and $\mathbf{f}_{:,d}^r$ denotes the $d $th column of $\mathbf{F}^r $.
The independence assumption embedded in Equation~\eqref{eq:prior1} and Equation~\eqref{eq:prior2} is a direct consequence of the dual formulation in PPCA. A detailed explanation of duality is provided in the Appendix~\ref{appendix_ppca}.

\subsubsection{Observed Data Model}
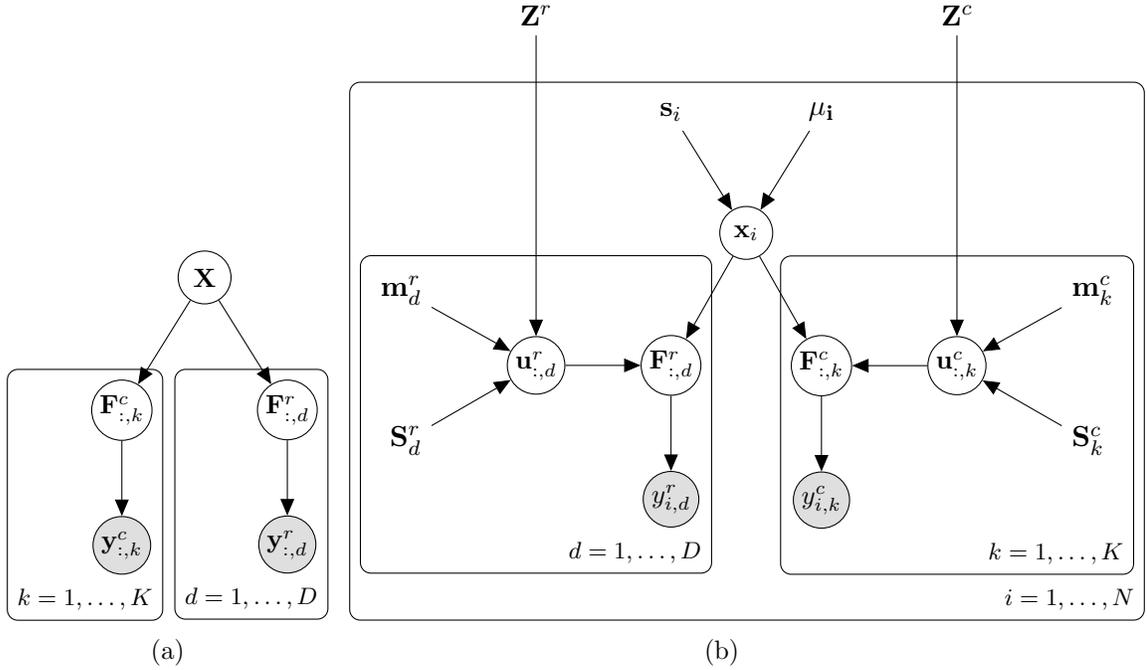
\begin{figure}
    \centering
    \begin{subfigure}{0.30\textwidth}
        \centering
        \begin{tikzpicture}

            \node[latent] (X) {$\mathbf{X}$};
            \node[latent, below=of X, xshift=-1.1cm] (Fs) {$\mathbf{F}^{c}_{:,k}$};
            \node[latent, below=of X, xshift=1.1cm] (Fn) {$\mathbf{F}^{r}_{:,d}$};
            \node[obs, below=of Fs] (Ys) {$\mathbf{y}^{c}_{:,k}$};
            \node[obs, below=of Fn] (Yn) {$\mathbf{y}^{r}_{:,d}$};

            \edge {X} {Fs} ; %
            \edge {X} {Fn} ; %
            \edge {Fn} {Yn} ; %
            \edge {Fs} {Ys} ; %
        
            \plate {S-plate} {(Ys)(Fs)} {$k=1,\ldots,K$} ;
            \plate {N-plate} {(Yn)(Fn)} {$ d=1,\ldots,D $} ;
        
        \end{tikzpicture}
        \caption{ }
    \end{subfigure}
    \begin{subfigure}{0.65\textwidth}
        \centering
        \begin{tikzpicture}
            \node[latent] (X) {$\mathbf{x}_i$};
            \node[latent, below=of X, xshift=1cm] (Fs) {$\mathbf{F}^{c}_{:,k}$};
            \node[latent, below=of X, xshift=-1cm] (Fn) {$\mathbf{F}^{r}_{:,d}$};
            \node[obs, below=of Fs] (Ys) {$y^{c}_{i,k}$};
            \node[obs, below=of Fn] (Yn) {$y^{r}_{i,d}$};
            \node[latent, right=of Fs] (Us) {$\mathbf{u}^{c}_{:,k}$};
            \node[latent, left=of Fn] (Un) {$\mathbf{u}^{r}_{:,d}$};
            \node[above=4cm of Un] (Zn) {$\mathbf{Z}^{r}$};
            \node[above=4cm of Us] (Zs) {$\mathbf{Z}^{c}$};
            
            \node[above=of X, xshift=1cm] (mu) {$\mathbf{\mu_{i}}$};
            \node[above=of X, xshift=-1cm] (s) {$\mathbf{s}_{i}$};

            \node[right=of Us, yshift=1cm] (m_k) {$\mathbf{m}^{c}_{k}$};
            \node[right=of Us, yshift=-1cm] (s_k) {$\mathbf{S}^{c}_{k}$};

            \node[left=of Un, yshift=1cm] (m_d) {$\mathbf{m}^{r}_{d}$};
            \node[left=of Un, yshift=-1cm] (s_d) {$\mathbf{S}^{r}_{d}$};

            \edge {X} {Fs} ; %
            \edge {X} {Fn} ; %
            \edge {Fn} {Yn} ; %
            \edge {Fs} {Ys} ; %
            \edge {Us} {Fs} ; %
            \edge {Un} {Fn} ; %
            \edge {Zn} {Un} ; %
            \edge {Zs} {Us} ; %

            \edge {mu} {X} ; %
            \edge {s} {X} ; %

            \edge {m_d} {Un} ; %
            \edge {s_d} {Un} ; %
            \edge {m_k} {Us} ; %
            \edge {s_k} {Us} ; %
        
            \plate {S-plate} {(Ys)(Fs)(Us)(s_k)(m_k)} {$k=1,\ldots,K$} ;
            \plate {d-plate} {(Yn)(Fn)(Un)(s_d)(m_d)} {$ d=1,\ldots, D$} ;
            \plate {N-plate} {(d-plate)(S-plate)(X)(mu)(s)} {$ i=1,\ldots,N$} ;
        \end{tikzpicture}
        \caption{ }
    \end{subfigure}
    \caption{Graphical Models Depicting LDGD. (a) Exact inference, (b) Variational inference}
    \label{fig:diagram}
\end{figure}

LDGD delineates the relationship between the features and the observed data. This relationship is modeled by:
\begin{align}
    p(\mathbf{Y}^r \mid \mathbf{F}^r) &= \prod_{i=1}^{N} \prod_{d=1}^{D} \mathcal{N}(y_{i,d}^r \mid f_{d}^{r}(\mathbf{x}_i), \sigma_d^2) \\
    p(\mathbf{Y}^c \mid \mathbf{F}^c) &= \prod_{i=1}^{N} \prod_{k=1}^{K} \text{Bernoulli}(y_{i,k}^c \mid g(f_{k}^{c}(\mathbf{x}_i))),
    \label{eq:gp_prior}
\end{align}
where $\sigma_d^2$ is the variance of the Gaussian noise and $g(\cdot)$ is a squash function, such as the Sigmoid or Probit, ensuring the output represents a probability. For the $i$th element in the $d $th column of $\mathbf{F}^r $, we define $ f^r_{i,d} \triangleq f^r_d(\mathbf{x}_i)$. Similarly, for the $i$th element in the $k$th column of $\mathbf{F}^{c}$, we define $ f^c_{i,k} \triangleq f^c_k(\mathbf{x}_i)$.
Figure~\ref{fig:diagram} shows the graphical model behind the LDGD framework.

\subsubsection{Marginal Likelihood Lower Bound}
\label{section:ELBO}
\label{subsec:marginal_likelihood_lower_bound}
Equation sets \eqref{eq:BGPLVM_prior} to \eqref{eq:gp_prior} define the LDGD model. We need to find the marginal distribution of $Y^{r}$ and $Y^{c}$ to fit the model to the dataset. The marginal distribution is defined by: 
\begin{equation}
    p(\mathbf{Y}^r, \mathbf{Y}^c) = \int p(\mathbf{Y}^r, \mathbf{Y}^c, \mathbf{F}^r, \mathbf{F}^c, \mathbf{X}) \, d\mathbf{F}^r \, d\mathbf{F}^c \, d\mathbf{X}.
    \label{eq:marginalization_evidence}
\end{equation}
Similar to Equation~\eqref{eq:sgpr_evidence}, this integral has no closed-form solution. Additionally, calculating the integral numerically is impractical due to the high dimensionality of the latent elements over which the integral is taken. Given there is no solution to find this integral, maximizing the evidence by adjusting its free parameter becomes impossible. Instead, we derive a lower bound for the marginal likelihood, similar to what we showed in Equation~\eqref{eq:sgpr_elbo}. To achieve this, we will be using inducing points and variational distributions. As we show later, our solution will have a modest computational cost and can be applied to high-dimensional datasets. Besides, the model training can be done through batches of datasets, making it appropriate for iterative training.

\paragraph{Inducing Points and Variational Distributions}
In Section~\ref{section:gpr}, we have explained the utilization of inducing points for GP regression. Similarly, in LDGD, we apply the same concept, with the distinction that we now need to address both discrete and continuous observation processes. Attributes of these processes impose different sensitivities to the choice of inducing points; we pick two sets of inducing points, one for the discrete and one for the continuous observations.
For the label data in the classification component, we define the inducing points and variables as
\begin{equation}
    p(\mathbf{U}^{c} \mid \mathbf{Z}^{c}) = \prod_{k=1}^{K} \mathcal{N}(\mathbf{u}_{:,k}^{c} \mid \mathbf{0}, k^{c} (\mathbf{Z}^c, \mathbf{Z}^{c})),
    \label{eq:inducing_variables_cls}
\end{equation}
where $\mathbf{U}^{c} \equiv \left[\mathbf{u}_{:,1}^{c}, \ldots, \mathbf{u}_{:,K}^{c}\right]$, $\mathbf{u}_{:,k}^{c} \in \mathbb{R}^M$, and $\mathbf{K}_{M_cM_c}^{c} \triangleq k^{c} (\mathbf{Z}^c, \mathbf{Z}^{c})$. Here, $\mathbf{Z}^{c}$ represents the set with $M_c$ inducing points associated with the $\mathbf{F}^{c}$.
Analogously, for the continuous measurement of the regression component, we have:
\begin{equation}
    p(\mathbf{U}^{r} \mid \mathbf{Z}^{r}) = \prod_{d=1}^{D} \mathcal{N}(\mathbf{u}_{:,d}^{r} \mid \mathbf{0}, k^{r} (\mathbf{Z}^{r}, \mathbf{Z}^{r})),
    \label{eq:inducing_variables_reg}
\end{equation}
where $\mathbf{U}^{r} \equiv \left[\mathbf{u}_{:,1}^{r}, \ldots, \mathbf{u}_{:,D}^{r}\right]$, $\mathbf{u}_{:,d}^{r} \in \mathbb{R}^M$, and $\mathbf{K}_{M_rM_r}^{r} \triangleq k^{r} (\mathbf{Z}^{r}, \mathbf{Z}^{r})$. Here, we use a new set of $M_r $ inducing points, $Z^{r}$, associated with the $\mathbf{F}^{r}$. The choice of two disjoint sets of inducing points addresses different sensitivities we have in the regression and classification. 

\paragraph{Doubly Stochastic Variational GP and Variational Posterior Distribution.}
\label{sec:doubly_stochastic}
The true posterior distribution is given by
\begin{equation}
    p(\mathbf{F}^r, \mathbf{F}^c, \mathbf{X} \mid \mathbf{Y}^r, \mathbf{Y}^c) 
    = \frac{p(\mathbf{F}^r, \mathbf{F}^c, \mathbf{X}, \mathbf{Y}^r, \mathbf{Y}^c)}
    {p(\mathbf{Y}^r, \mathbf{Y}^c)}.
\end{equation}
Calculating the denominator involves the computation of Equation~\eqref{eq:marginalization_evidence}, which is intractable. We propose variational distributions as a part of our approach to approximate the posterior distribution in LDGD. Our methodology incorporates concepts from the doubly stochastic variational Gaussian process (DSVGP) outlined by \cite{salimbeni2017doubly}, and integrates the scalable variational GP framework for classification proposed by \cite{hensman2015scalable}. This integration is achieved through the utilization of two distinct sets of inducing variables, as delineated in Equations~\eqref{eq:inducing_variables_cls} and \eqref{eq:inducing_variables_reg}, which are applied respectively for classification and regression tasks. Training involves maximizing the marginal likelihood (evidence), and our method provide a means to find a lower bound for the otherwise intractable marginalization process. To achieve this lower bound, we begin by marginalizing the joint distribution of LDGD, taking into account the variational inducing points introduced: 
\begin{align}
\label{eq:ldgd_evidence}
    p(\mathbf{Y}^{r}, \mathbf{Y}^{c}) 
    =&\int
        p(\mathbf{Y}^{r}, \mathbf{Y}^{c}, \mathbf{F}^{r}, \mathbf{F}^{c}, \mathbf{U}^{r}, \mathbf{U}^{c}, \mathbf{X}) \, 
        d\mathbf{F}^{r} \, d\mathbf{F}^{c} \, d\mathbf{U}^{r} \, d\mathbf{U}^{c} \, d\mathbf{X}.
\end{align}
The joint distribution of LDGD, given the observed data $\mathbf{Y}^r$ and $\mathbf{Y}^c$, the function values $\mathbf{F}^r$ and $\mathbf{F}^c$, the inducing variables $\mathbf{U}^{r}$ and $\mathbf{U}^{c}$, and the latent features $\mathbf{X}$, can be factorized as:
\begin{align}
\label{eq:ldgd_joint_dist}
    p(\mathbf{F}^r, \mathbf{F}^c, \mathbf{U}^{r}, \mathbf{U}^c, \mathbf{X}, \mathbf{Y}^r, \mathbf{Y}^c) 
    =& p(\mathbf{Y}^{c} \mid \mathbf{F}^{c})
    p(\mathbf{F}^{c}, \mathbf{U}^{c} \mid \mathbf{X})
    p(\mathbf{Y}^{r} \mid \mathbf{F}^{r})
    p(\mathbf{F}^{r}, \mathbf{U}^{r} \mid \mathbf{X})
    p(\mathbf{X}).
\end{align}
Solving this integral is not straightforward. A common technique used in variational inference is multiplying and dividing the terms inside the integral by a proposal distribution, which is an approximation of the joint posterior. The joint posterior \(p(\mathbf{F}^{r}, \mathbf{F}^{c}, \mathbf{U}^{r}, \mathbf{U}^{c}, \mathbf{X} \mid \mathbf{Y}^{r}, \mathbf{Y}^{c})\) can be factorized into three separate posteriors:
\begin{equation}
\label{eq:posterior_with_induce}
    p(\mathbf{F}^{r}, \mathbf{F}^{c}, \mathbf{U}^{r}, \mathbf{U}^{c}, \mathbf{X} \mid \mathbf{Y}^{r}, \mathbf{Y}^{c}) = 
        p(\mathbf{F}^{c}, \mathbf{U}^{c} \mid \mathbf{Y}^{c}, \mathbf{X}) 
        p(\mathbf{F}^{r}, \mathbf{U}^{r} \mid \mathbf{Y}^{r}, \mathbf{X}) 
        p(\mathbf{X} \mid \mathbf{Y}^{r}, \mathbf{Y}^{c}).
\end{equation}
where $p(\mathbf{F}^{c}, \mathbf{U}^{c} \mid \mathbf{Y}^{c}, \mathbf{X})$ and $p(\mathbf{F}^{r}, \mathbf{U}^{r} \mid \mathbf{Y}^{r}, \mathbf{X})$ are the joint posterior of the Gaussian process function values and their inducing variables for regression and classification respectively. $p(\mathbf{X} \mid \mathbf{Y}^{r}, \mathbf{Y}^{c})$ shows the posterior over the latent features. To derive a tractable solution using posterior, we approximate posterior $p(\mathbf{X}\mid \mathbf{Y}^{r}, \mathbf{Y}^{c})$ with variational distribution $q_{\phi}(\mathbf{X})$, $p(\mathbf{F}^{c}, \mathbf{U}^{c} \mid \mathbf{Y}^{c}, \mathbf{X})$ with $q(\mathbf{F}^c, \mathbf{U}^{c})$ and $p(\mathbf{F}^{r}, \mathbf{U}^{r} \mid \mathbf{Y}^{r}, \mathbf{X})$ with $q(\mathbf{F}^{r}, \mathbf{U}^{r})$:
\begin{align}
    q_{\phi}(\mathbf{X}) 
    &= \prod_{i=1}^{N} \mathcal{N}(\mathbf{x}_i \mid \boldsymbol{\mu}_i, \mathbf{s}_i \mathbb{I}_Q), \label{eq:x_variational_aprox}\\
    q(\mathbf{F}^c, \mathbf{U}^{c}) 
    &= p(\mathbf{F}^c \mid \mathbf{U}^c, \mathbf{X}) q_{\lambda}(\mathbf{U}^{c}), \quad q_{\lambda}(\mathbf{U}^{c}) 
    = \prod_{k=1}^{K} \mathcal{N}(\mathbf{u}_k^{c} \mid \mathbf{m}_k^c, \mathbf{S}_k^{c}), \\
    q(\mathbf{F}^r, \mathbf{U}^{r}) 
    &= p(\mathbf{F}^r \mid \mathbf{U}^{r}, \mathbf{X}) 
    q_{\gamma}(\mathbf{U}^{r}), \quad q_{\gamma}(\mathbf{U}^{r}) = \prod_{d=1}^{D} \mathcal{N}(\mathbf{u}_d^{r} \mid \mathbf{m}_d^{r}, \mathbf{S}_d^{r}),
    \label{eq:variational_distributions}
\end{align}
where $\phi = \{\boldsymbol{\mu}_i, \mathbf{s}_i\}_{i=1}^{N}$, $\lambda = \{\mathbf{m}_k^c, \mathbf{S}_k^{c}\}_{k=1}^{K}$,and $\gamma = \{\mathbf{m}_d^{r}, \mathbf{S}_d^{r}\}_{d=1}^{D}$ are the variational parameters. Using the variational approximations, the joint posteriors can be expressed as:
\begin{equation}
    \label{eq:variational_posterior}
    p(\mathbf{F}^r, \mathbf{F}^c, \mathbf{U}^{r}, \mathbf{U}^c, \mathbf{X} \mid \mathbf{Y}^r, \mathbf{Y}^c) 
    \approx q(\mathbf{F}^{r}, \mathbf{F}^{c}, \mathbf{U}^{r}, \mathbf{U}^{c}, \mathbf{X})
    \equiv q(\mathbf{F}^c, \mathbf{U}^{c}) q(\mathbf{F}^r, \mathbf{U}^{r}) q_{\phi}(\mathbf{X}).
\end{equation}
Employing the approximated posterior, we reformulate the evidence introduced in Equation~\eqref{eq:ldgd_evidence} and apply Jensen's inequality on the logarithm of the evidence to establish the evidence lower-bound (\(\text{ELBO}\)):
\begin{align*}
    \log p(\mathbf{Y}^{r}, \mathbf{Y}^{c}) 
    \geq&
    E_{q(\mathbf{F}^{r}, \mathbf{F}^{c}, \mathbf{U}^{r}, \mathbf{U}^{c}, \mathbf{X})}\left[
        \log \frac{
            p(\mathbf{Y}^{c} \mid \mathbf{F}^{c})p(\mathbf{U}^{c})
            p(\mathbf{Y}^{r} \mid \mathbf{F}^{r})p(\mathbf{U}^{r})
            p(\mathbf{X})
        }{
            q_{\phi}(\mathbf{X})
            q_{\lambda}(\mathbf{U}^{c})
            q_{\gamma}(\mathbf{U}^{r})
        }
    \right] \triangleq \text{ELBO}.
\end{align*}
We can further breakdown this ELBO in 5 terms
\begin{align}
    \text{ELBO} &\equiv \text{ELL}^{\text{reg}} + \text{ELL}^{\text{cls}} - \text{KL}_{u}^{c}- \text{KL}_{u}^{r}- \text{KL}_{X} ,
\label{eq:elbo_all}
\end{align}
where the $\text{ELL}^{reg}$ and $\text{ELL}^{cls}$ are expected log-likelihood (ELL) for regression and classification paths, respectively, and are defined by:
\begin{align*}
    \text{ELL}^{\text{reg}} &\equiv 
    E_{q_{\phi}(\mathbf{X})} 
    \left[
        E_{p(\mathbf{F}^{r} \mid \mathbf{U}^{r}, \mathbf{X}) 
        q_{\gamma}(\mathbf{U}^{r})}
        \left[\log p(\mathbf{Y}^{r} \mid \mathbf{F}^{r})\right]
    \right],  \\
    \text{ELL}^{\text{cls}} &\equiv 
    E_{q_{\phi}(\mathbf{X})} 
    \left[
        E_{p(\mathbf{F}^{c} \mid \mathbf{U}^{c}, \mathbf{X}) 
        q_{\lambda}(\mathbf{U}^{c})}
        \left[\log p(\mathbf{Y}^{c} \mid \mathbf{F}^{c})\right] 
    \right].
\end{align*}
The KL terms are defined as:
\begin{align*}
    &\text{KL}_{u}^{c} \equiv \textit{KL}(q_{\lambda}(\mathbf{U}^{c}) \parallel p(\mathbf{U}^{c})), \\
    &\text{KL}_{u}^{r} \equiv  \textit{KL}(q_{\gamma}(\mathbf{U}^{r}) \parallel p(\mathbf{U}^{r})), \\
    &\text{KL}_{X} \equiv \textit{KL}(q_{\phi}(\mathbf{X}) \parallel p(\mathbf{X})).
\end{align*}
 $\text{ELL}^{reg}$ and $\text{ELL}^{cls}$ contribute to data fitting, whereas $\text{KL}_{u}^{c}$, $\text{KL}_{u}^{r}$, and $\text{KL}_{X}$ terms serve as regularizers and help prevent overfitting. Minimizing the KL divergence between the posterior defined in Equation~\eqref{eq:posterior_with_induce} and its variational estimation defined in Equation~\eqref{eq:variational_posterior} is equivalent to maximizing the $\text{ELBO}$. A detailed explanation is provided in Appendix~\ref{apendix_LDGD_elbo}. A closed form for this lower bound is desirable for optimizing the variational parameters, inducing points, and kernel parameters. While KL terms have a closed form due to the Gaussian distribution characteristics, $\text{ELL}^{cls}$ and $\text{ELL}^{reg}$ do not have a closed form in general. The details of calculating these two terms are explained in the subsequent sections.

\subsubsection{Classification Expected Log-Likelihood ($\text{ELL}^{\text{cls}}$)}
In this section, we delve into the detailed analysis of $\text{ELL}^{\text{cls}}$, defined as
\begin{align}
    \text{ELL}^{\text{cls}} &= E_{q_{\phi}(\mathbf{X})} 
    \left[
        E_{p(\mathbf{F}^{c} \mid \mathbf{U}^{c}, \mathbf{X}) 
        q_{\lambda}(\mathbf{U}^{c})}
        \left[\log p(\mathbf{Y}^{c} \mid \mathbf{F}^{c})\right] 
    \right] \notag \\
    &= \int 
    q_{\phi}(\mathbf{X}) 
    p(\mathbf{F}^c \mid \mathbf{U}^c, \mathbf{X}) 
    q_{\lambda}(\mathbf{U}^{c}) \sum_{i,k} 
    \log p(y_{i,k}^c \mid f_k^{c}(\mathbf{x}_i)) 
    \, d\mathbf{F}^c \, d\mathbf{U}^{c} \, d\mathbf{X} \notag\\
    &= \sum_{i,k} \int 
    q_{\phi}(\mathbf{x}_i) p(f_k^{c} \mid \mathbf{u}_k^c, \mathbf{x}_i) 
    q_{\lambda}(\mathbf{u}_k^{c}) \log p(y_{i,k}^c \mid f_k^{c}(\mathbf{x}_i)) \, 
    df_k^{c} \, d\mathbf{u}_k^{c} \, d\mathbf{x}_i \notag \\
    &= \sum_{i,k} ELL_{i,k}^{\text{cls}}.
    \label{eq:ell_cls1}
\end{align}

We rewrite the $ELL_{i,k}^{\text{cls}}$:
\begin{align*}
    ELL_{i,k}^{\text{cls}} &= \int q_{\phi}(\mathbf{x}_i) p(f_k^{c} \mid \mathbf{u}_k^c, \mathbf{x}_i) q_{\lambda}(\mathbf{u}_k^{c}) \log p(y_{i,k}^c \mid f_k^{c}(\mathbf{x}_i)) \, df_k^{c} \, d\mathbf{u}_k^{c} \, d\mathbf{x}_i \notag \\
    &= \int q_{\phi}(\mathbf{x}_i) q_{\lambda}(f_k^{c} \mid \mathbf{x}_i) \log p(y_{i,k}^c \mid f_k^{c}(\mathbf{x}_i)) \, df_k^{c} \, d\mathbf{x}_i \notag \\
    &= E_{q_{\phi}(\mathbf{x}_i) q_{\lambda}(f_k^{c} \mid \mathbf{x}_i)} \left[ \log p(y_{i,k}^c \mid f_k^{c}(\mathbf{x}_i)) \right],
\end{align*}
where we define the predictive distribution as:
\begin{align*}
    q_{\lambda}(f_k^{c} \mid \mathbf{x}_i) 
    &\triangleq \int p(f_k^{c} \mid u_{k}^c, \mathbf{x}_i) q_{\lambda}(\mathbf{u}_k^{c}) \, d\mathbf{u}_k^{c} = \mathcal{N}\left( \boldsymbol{\mu}_{f_k^{c}}(\mathbf{x}_i), \boldsymbol{\Sigma}_{f_{k}^{c}}(\mathbf{x}_i) \right)\\
\boldsymbol{\mu}_{f_k^{c}}(\mathbf{x}_i) &=
\mathbf{k}_{iM_c}^{c} 
(\mathbf{K}_{M_cM_c}^{c})^{-1} 
\mathbf{m}_{k}^{c} \\
\boldsymbol{\Sigma}_{f_{k}^{c}}(\mathbf{x}_i)&=
k_{ii}^{c} + 
\mathbf{k}_{iM_c}^{c} 
(\mathbf{K}_{M_cM_c}^{c})^{-1} 
(\mathbf{S}_d^{c} - 
\mathbf{K}_{M_cM_c}^{c}) 
(\mathbf{K}_{M_cM_c}^{c})^{-1} 
\mathbf{K}_{M_ci}^{c},
\end{align*}
with $\mathbf{k}^{c}_{iM_c} = k^{c}(\mathbf{x}_i, \mathbf{Z}^{c})$ being the $i$th row of $K^{c}_{NM_c}$ and $k^c_{ii} = k^{c}(\mathbf{x}_i, \mathbf{x}_i)$. The details are provided in Appendix~\ref{appendix_predictive_dist}. For ease of use, we define  $B(\mathbf{x}_i) \triangleq E_{q_{\gamma}(\mathbf{f}_{k}^{c} \mid \mathbf{x}_i)} \left[\log p(y_{i,k}^c \mid f_{k}^{c}(\mathbf{x}_i)) \right]$. For the Bernoulli likelihood with Probit likelihood function, the inner expectation $B(\mathbf{x}_i)$ is:
\begin{align*}
B(\mathbf{x}_i) &= 
    E_{q_{\gamma}(\mathbf{f}_{k}^{c} \mid \mathbf{x}_i)} 
\left[\log p(y_{i,k}^c \mid f_{k}^{c}(\mathbf{x}_i)) \right] \notag \\
&= \int 
    q_{\gamma}(\mathbf{f}_{k}^{c} \mid \mathbf{x}_i) 
    \log \Phi((2y_{i,k}^c - 1) f_{k}^{c}(\mathbf{x}_i))\, 
    df_{k}^{c} \notag \\
    &= \int 
    \mathcal{N}\left(\mathbf{f}_{k}^{c} \mid \boldsymbol{\mu}_{f_{k}^{c}}
    (\mathbf{x}_i), \boldsymbol{\Sigma}_{f_{k}^{c}}(\mathbf{x}_i) \right) 
    \log \Phi((2y_{i,k}^c - 1) f_{k}^{c}(\mathbf{x}_i)) 
    \, d\mathbf{f}_{k}^{c} \notag \\
&= \int 
    \frac{1}{\sqrt{2\pi \boldsymbol{\Sigma}_{f_{k}^{c}}(\mathbf{x}_i)}} 
    \exp\left(
        -\frac{
            (\mathbf{f}_{k}^{c} 
            - \boldsymbol{\mu}_{f_{k}^{c}}(\mathbf{x}_i))^2
        }{
            2\boldsymbol{\Sigma}_{f_{k}^{c}}(\mathbf{x}_i)
        }\right) 
    \log \Phi((2y_{i,k}^c - 1) f_{k}^{c}(\mathbf{x}_i)) \, 
    d\mathbf{f}_{k}^{c} .
\end{align*}
To use Gauss-Hermite quadrature, we need to transform this integral into the standard form of $\int_{-\infty}^{\infty} e^{-x^2} g(x) \, dx $. We can perform a change of variable $\mathbf{z} = 
\frac{\mathbf{f}_{k}^{c} - \boldsymbol{\mu}_{f_{k}^{c}}(\mathbf{x}_i)}
{\sqrt{2\boldsymbol{\Sigma}_{\mathbf{f}_{k}^{c}}(\mathbf{x}_i)}}$ so that $\mathbf{f}_{k}^{c} = \mathbf{z}\sqrt{2\boldsymbol{\Sigma}_{\mathbf{f}_{k}^{c}}(\mathbf{x}_i)} + \boldsymbol{\mu}_{f_{k}^{c}}(\mathbf{x}_i)$ and $d\mathbf{f}_{k}^{c} = \sqrt{2\boldsymbol{\Sigma}_{f_{k}^{c}}(\mathbf{x}_i)} \, dz$. The integral becomes:
\begin{equation*}
B(\mathbf{x}_i) = 
\int_{-\infty}^{\infty} \frac{e^{-z^2}}{\sqrt{\pi}} 
\log \Phi
\left(
    (2y_{i,k}^c - 1)
    \left(
        z\sqrt{2\boldsymbol{\Sigma}_{f_{k}^{c}}(\mathbf{x}_i)} 
        + \boldsymbol{\mu}_{f_{k}^{c}}(\mathbf{x}_i)
    \right)
\right) 
\sqrt{2\boldsymbol{\Sigma}_{f_{k}^{c}}(\mathbf{x}_i)} 
\, dz.
\end{equation*}
Now, the integral is in a form suitable for Gauss-Hermite quadrature, where 
\begin{equation*}  \sqrt{\boldsymbol{\Sigma}_{\mathbf{f}_{k}^{c}}(\mathbf{x}_i)} 
\log \Phi((2y_{i,k}^c - 1)(\mathbf{z}
\sqrt{2\boldsymbol{\Sigma}_{f_{k}^{c}}(\mathbf{x}_i)} 
+ \boldsymbol{\mu}_{f_{k}^{c}}(\mathbf{x}_i))).
\end{equation*}
Let $z^{(j)}_{i}$ and $w^{(j)}$ denote the nodes and weights of the Gauss-Hermite quadrature. The integral $B(\mathbf{x}_i)$ can be approximated as:
\begin{equation*}
B(\mathbf{x}_i) \approx \frac{1}{\sqrt{\pi}} \sum_{l=1}^{L} w^{(l)} \log \Phi((2y_{i,k}^c - 1)(z^{(l)}_{i}\sqrt{2\boldsymbol{\Sigma}_{f_{k}^{c}}(\mathbf{x}_i)} + \boldsymbol{\mu}_{f_{k}^{c}}(\mathbf{x}_i))) \sqrt{2\boldsymbol{\Sigma}_{f_{k}^{c}}(\mathbf{x}_i)},
\end{equation*}
where $L$ is the number of quadrature points. The $z^{(l)}_{i}$ are the roots of the physicists' version of the Hermite polynomial $H_n(z_{i})$, and the associated weights $w^{(l)}$ are given by
\begin{equation*}
    w^{(l)} = \frac{2^{L-1} L! \sqrt{\pi}}{L^2 [H_{L-1}(z_{i})]^2}.
\end{equation*}
This approximation allows for the numerical evaluation of the integral using the Gauss-Hermite quadrature method \citep{abramowitz1988handbook}. to calculate the $E_{q_{\phi}(\mathbf{x}_i)} \left[ B(\mathbf{x}_i) \right]$ we can draw$J$ sample from $\mathbf{x}_i^{(j)} \sim q_{\phi}(\mathbf{x}_i)$ and estimate the expectation. 
\begin{align}
\label{eq:ell_cls_estimate}
    E_{q_{\phi}(\mathbf{x}_i)} \left[ B(\mathbf{x}_i) \right]
    \approx& 
    \frac{1}{\sqrt{\pi}} \sum_{l=1}^{L}\sum_{j=1}^{J} w^{(l)} 
    \log \Phi((2y_{i,k}^c - 1)(z^{(l)}_{i}
    \sqrt{2\boldsymbol{\Sigma}_{f_{k}^{c}}(\mathbf{x}^{(j)}_i)} +\\
    &\boldsymbol{\mu}_{f_{k}^{c}}(\mathbf{x}^{(j)}_i))) 
    \sqrt{2\boldsymbol{\Sigma}_{f_{k}^{c}}(\mathbf{x}^{(j)}_i)}.
\end{align}

\subsubsection{Regression Expected Log Likelihood ($ELL^{\text{reg}}$)}

In this section, we address the regression component of our model. With a similar calculation we have the:
\begin{align*}
    \text{ELL}^{\text{reg}} &= E_{q_{\phi}(\mathbf{X})} 
    \left[
        E_{p(\mathbf{F}^{r} \mid \mathbf{U}^{r}, \mathbf{X}) 
        q_{\lambda}(\mathbf{U}^{r})}
        \left[\log p(\mathbf{Y}^{r} \mid \mathbf{F}^{r})\right] 
    \right] \notag \\
    &= \int 
    q_{\phi}(\mathbf{X}) 
    p(\mathbf{F}^{r} \mid \mathbf{U}^{r}, \mathbf{X}) 
    q_{\gamma}(\mathbf{U}^{r}) \sum_{i,d} 
    \log p(y_{i,d}^{r} \mid f_{d}^{r}(\mathbf{x}_i)) 
    \, d\mathbf{F}^{r} \, d\mathbf{U}^{r} \, d\mathbf{X} \notag\\
    &= \sum_{i,d} \int 
    q_{\phi}(\mathbf{x}_r) p(\mathbf{f}_{d}^{r} \mid \mathbf{u}_d^r, \mathbf{x}_i) 
    q_{\gamma}(\mathbf{u}_d^{r}) 
    \log p(y_{i,d}^r \mid f_d^{r}(\mathbf{x}_i)) \, 
    d\mathbf{f}_{d}^{r} \, d\mathbf{u}_d^{r} \, d\mathbf{x}_i \notag \\
    &= \sum_{i,d} ELL_{i,d}^{\text{reg}}.
\end{align*}
Here are the computation details for calculating the ELL:
\begin{align}
ELL_{i,d}^{\text{reg}} &= 
\int 
    q_\phi (\mathbf{x}_i) p(\mathbf{f}_{d}^{r} \mid \mathbf{u}_{d}^{r}, \mathbf{x}_i) 
    q_\gamma (\mathbf{u}_{d}^{r}) 
    \log p(y_{i,d}^{r} \mid f_{d}^{r} (\mathbf{x}_i)) \, 
    d\mathbf{f}_{d}^{r} \, d\mathbf{u}_{d}^{r} \, d\mathbf{x}_i \notag \\
&= \int 
    q_\phi (\mathbf{x}_i)  
    \int 
        q_\gamma (\mathbf{u}_{d}^{r})  p(\mathbf{f}_{d}^{r} \mid \mathbf{u}_{d}^{r}, \mathbf{x}_i) \, d\mathbf{u}_{d}^{r}  \, 
        \log p(y_{i,d}^{r} \mid f_{d}^{r} (\mathbf{x}_i)) 
        d\mathbf{f}_{d}^{r} \, 
    d\mathbf{x}_i \notag \\
&= \int 
    q_\phi (\mathbf{x}_i) q_{\gamma}(\mathbf{f}_{d}^{r} \mid \mathbf{x}_i) 
    \log p(y_{i,d}^{r} \mid f_{d}^{r} (\mathbf{x}_i))  \, 
    d\mathbf{f}_{d}^{r} \, d\mathbf{x}_i \notag \\
&= E_{q_{\phi}(\mathbf{x}_i)} 
\left[ 
    E_{q_{\gamma}(\mathbf{f}_{d}^{r} \mid \mathbf{x}_i)} 
    \left[
        \log p(y_{i,d}^{r} \mid f_{d}^{r} (\mathbf{x}_i)) 
    \right] 
\right] \notag \\
&= E_{q_{\phi}(\mathbf{x}_i)}
    \left[A(\mathbf{x}_i)\right].
\label{eq:ell_reg1}
\end{align}
We already know that $y_{i,d}^{r} = f_{d}^{r} (\mathbf{x}_i) + \epsilon_d$, therefore the internal term has a Gaussian distribution 
\begin{equation*}
    p(y_{i,d}^{r} \mid f_{d}^{r} (\mathbf{x}_i))= \mathcal{N}(f_{d}^{r} (\mathbf{x}_i), \boldsymbol{\Sigma}_d^2).
\end{equation*}
from Appendix~\ref{appendix_predictive_dist} we know the predictive distribution. 
\begin{align*}
    q_{\gamma}(f_{d}^{r} \mid \mathbf{x}_i) &= \mathcal{N}\left(f_{d}^{r} \mid \boldsymbol{\mu}_{\mathbf{f}_{d}^{r}}(\mathbf{x}_i), \boldsymbol{\Sigma}_{\mathbf{f}_{d}^{r}}(\mathbf{x}_i) \right) \\
    \boldsymbol{\mu}_{\mathbf{f}_{d}^{r}}(\mathbf{x}_i) &= \mathbf{k}_{iM_r}^{r} (\mathbf{K}_{M_rM_r}^{r})^{-1} \mathbf{m}^r_d \\
    \boldsymbol{\Sigma}_{\mathbf{f}_{d}^{r}}(\mathbf{x}_i) &=
k_{ii}^{r} + 
\mathbf{k}_{iM_r}^{r} 
(\mathbf{K}_{M_rM_r}^{r})^{-1} 
(\mathbf{S}_{d}^{r} - 
\mathbf{K}_{M_rM_r}^{r}) 
(\mathbf{K}_{M_rM_r}^{r})^{-1} 
\mathbf{k}_{M_ri}^{r},
\end{align*}
where $\mathbf{k}^r_{iM_{r}} = k^{r}(\mathbf{x}_i, \mathbf{Z}^{r})$, $k_{ii}^{r} = k^{r}(\mathbf{x}_i, \mathbf{x}_i)$, and $K_{M_rM_r}^{r}$ is the covariance matrix evaluated at inducing points.

\begin{align*}
A(\mathbf{x}_i) 
&= E_{q_{\gamma}(f_{d}^{r} \mid \mathbf{x}_i)} 
\left[\log p(y_{i,d}^{r} \mid f_{d}^{r}(\mathbf{x}_i)) \right] \notag \\
&= \int 
    q_{\gamma}(f_{d}^{r} \mid \mathbf{x}_i) 
    \log p(y_{i,d}^{r} \mid f_{d}^{r} (\mathbf{x}_i))
    \, df_{d}^{r} \notag \\
&= -0.5 \left[ 
    \log 2\pi 
    + \log \boldsymbol{\Sigma}_d^2 +\frac{(y_{i,d}^{r}
    -\boldsymbol{\mu}_{\mathbf{f}_{d}^{r}}(\mathbf{x}_i))^2
    +\boldsymbol{\Sigma}_{\mathbf{f}_{d}^{r}}(\mathbf{x}_i)}{\boldsymbol{\Sigma}_d^2} .
    \right].
\end{align*}

We can easily calculate this by sampling.
\begin{align}
\label{eq:ell_reg_estimate}
ELL_{i,d}^{\text{reg}} &= E_{q_{\phi}(\mathbf{x}_i)} \left[ E_{q_{\gamma}(f_{d}^{r} \mid \mathbf{x}_i)} \left[\log p(y_{i,d}^{r} \mid f_{d}^{r}(\mathbf{x}_i)) \right] \right] \notag \\
&= -\frac{1}{2J} \sum_{j=1}^{J} \left[ \log 2\pi + \log \boldsymbol{\Sigma}_d^2 +\frac{(y_{i,d}^{r}-\boldsymbol{\mu}_{\mathbf{f}_{d}^{r}}(x^{(j)}_{i}))^2+\boldsymbol{\Sigma}_{\mathbf{f}_{d}^{r}}(x^{(j)}_{i})}{\boldsymbol{\Sigma}_{d}^2} \right].
\end{align}
It is worth mentioning that for some certain kernel functions, finding the analytic form for the expectation introduced in Equation~\eqref{eq:ell_reg1} is possible for certain kernel functions. This concludes our analysis and approximation of the Regression expected log-likelihood for the model.

\subsubsection{Training procedure}
Learning occurs through optimizing the kernel hyperparameters, inducing points, and variational parameters by maximizing the ELBO as the objective function. 

\begin{equation*}
    \argmax_{\mathbf{\theta}, \mathbf{\phi}, \mathbf{\lambda}, \mathbf{\gamma}, \mathbf{Z}^r, \mathbf{Z}^c} \text{ELBO}
\end{equation*}
Where $\mathbf{\theta}$ represents the kernel hyperparameters, $\mathbf{\phi} = \{(\boldsymbol{\mu}_i, \mathbf{s}_i), i = 1, \ldots, N\}$ are the variational parameters for latent features, $\mathbf{\lambda} = \{(\mathbf{m}_k^c, \mathbf{S}_k^c), k = 1, \ldots, K\}$ and $\mathbf{\gamma} = \{(\mathbf{m}_d^r, \mathbf{S}_d^r), d = 1, \ldots, D\}$ are the variational parameters for inducing variables. In the previous section, we discussed how the expected log-likelihood for regression ($\text{ELL}^{reg}$) and classification ($\text{ELL}^{cls}$) can be broken down and calculated for each sample in the dataset. This formulation simplifies the calculations and enables batch training and sampling methods. To optimize the variational parameters for $q_{\phi}(\mathbf{X})$ during sampling, We generate samples $\mathbf{x}_i^{(j)} \sim q_{\phi}(\mathbf{x}_i)$ using the reparameterization technique \cite{}, where $\epsilon^{(j)} \sim \mathcal{N}(0, \mathbb{I}_Q)$ and $\mathbf{x}_i^{(j)} = \boldsymbol{\mu}_i + \mathbf{s}_i \odot \epsilon^{(j)}$. Based on Equation~\eqref{eq:elbo_all}, \eqref{eq:ell_reg1} , and \eqref{eq:ell_cls1}, the training procedure is simplified to this maximization problem:
\begin{align}
\argmax_{\mathbf{\theta}, \mathbf{\phi}, \mathbf{\lambda}, \mathbf{\gamma}, \mathbf{Z}^r, \mathbf{Z}^c} 
& \sum_{i,d} \text{ELL}_{i,d}^{\text{reg}} + \sum_{i,k} \text{ELL}_{i,k}^{\text{cls}} - \sum_{i=1}^{N} \text{KL}(q_\phi(\mathbf{x}_i) \parallel p(\mathbf{x}_i)) \notag \\
& - \sum_{k=1}^{K} \text{KL}(q_\lambda(\mathbf{u}_k^{c}) \parallel p(\mathbf{u}_k^{c} \mid \mathbf{Z}^{c})) 
- \sum_{d=1}^{D} \text{KL}(q_\gamma(\mathbf{u}_d^{r}) \parallel p(\mathbf{u}_d^{r} \mid \mathbf{Z}^{r}))
\end{align}

where the KL divergence terms have closed forms due to the conjugacy of the Gaussian distributions.$\text{ELL}_{i,d}^{\text{reg}}$ and $\text{ELL}_{i,k}^{\text{cls}}$ was introduced in Equation~\eqref{eq:ell_reg_estimate} and \eqref{eq:ell_cls_estimate}. The Algorithm~\ref{alg:train} details the training procedure. Detailed implementations of the predictive distribution, along with the calculations for the expected log-likelihood for regression and classification, are provided in Appendices \ref{appendix_predictive_dist} and \ref{appendix_ell}, respectively.

\begin{algorithm}[ht]
\caption{Training the Gaussian process Regression Model}
\begin{algorithmic}[1]

\Require 
\Statex Batch size ($B$), Learning rate ($lr$), Number of iterations ($N_{iter}$)
\Statex Training continuous measures $\mathbf{Y}^{r} \in \mathbb{R}^{N \times D}$
\Statex Training categorical observation or class labels $\mathbf{Y}^{c} \in \mathbb{R}^{N \times K}$

\Ensure Optimized model parameters $\{\theta^*, \psi^*, \phi^*, \lambda^*, \gamma^*, \mathbf{Z}^*\}$

\Procedure{Train Model}{$\mathbf{Y}^{r}, \mathbf{Y}^{c}, \mathbf{Z}^{c},\mathbf{Z}^{r}, lr, N_{iter}$}
    \State Initialize of inducing points $\mathbf{Z}^{c} \in \mathbb{R}^{M_{c} \times Q}$ and $\mathbf{Z}^{r} \in \mathbb{R}^{M_{r} \times Q}$
    \State Initialize kernel parameters $\mathbf{\psi}$ and $\mathbf{\theta}$
    \State Initialize likelihood noise variance $\boldsymbol{\Sigma}_y$
    \State Initialize  
   $\phi = \{(\boldsymbol{\mu}_i, \mathbf{s}_i) \mid \boldsymbol{\mu}_i \in \mathbb{R}^{Q}, \mathbf{s}_i \in \mathbb{R}^{Q}\}_{i=1}^{N}$
    for latent variables $\mathbf{X}$
    \State Initialize  
   $\lambda = \{(\mathbf{m}_d^r, \mathbf{S}_d^r) \mid \mathbf{m}_d^r \in \mathbb{R}^{M_r}, \mathbf{S}_d^r \in \mathbb{R}^{M_r \times M_r}\}_{d=1}^{D}$ for inducing variables $\mathbf{U}^{r}$
    
    \State Initialize  
   $\gamma = \{(\mathbf{m}_k^c, \mathbf{S}_k^c) \mid \mathbf{m}_k^c \in \mathbb{R}^{M_c}, \mathbf{S}_k^c \in \mathbb{R}^{M_c \times M_c}\}_{k=1}^{K}$ for inducing variables $\mathbf{U}^{c}$
    \State Initialize $ELBO$ to empty list
    \For{$ i=1$to$N_{iter}$}
        \State Sample $\boldsymbol{\epsilon}_i \sim \mathcal{N}(\mathbf{0}, \mathbb{I}_Q)$ for $ i = 1, \ldots, N$
        \State $\mathbf{x}_i \leftarrow \boldsymbol{\mu}_i + \sqrt{\mathbf{s}_i} \boldsymbol{\epsilon}_i$ for $ i = 1, \ldots, n $\Comment{Re-parameterization trick}
        \State $\mathbf{X} \leftarrow \{\mathbf{x}_1, \ldots, \mathbf{x}_N\}$
        \State Get a random batch $\mathbf{X}_b$
        \State Calculate $K^r_{NN}$, $\mathbf{K}_{M_rM_r}^{r}$, $\mathbf{K}_{M_rN}^{r}$, $\mathbf{K}^c_{NN}$, $\mathbf{K}^c_{M_cM_c}$, $\mathbf{K}^c_{M_cN}$
        
        \State $\boldsymbol{\mu}_{\mathbf{f}^r} , \boldsymbol{\Sigma}_{\mathbf{f}^r} \leftarrow$\Call{PredictiveDistribution}{$\mathbf{X}_b$, $\mathbf{K}^r_{NN}$, $\mathbf{K}^r_{M_rM_r}$, $\mathbf{K}^r_{M_rN}$, $\mathbf{m}^r $, $\mathbf{S}^r $}
        
        \State $\boldsymbol{\mu}_{\mathbf{f}^c} , \boldsymbol{\Sigma}_{\mathbf{f}^c} \leftarrow$\Call{PredictiveDistribution}{$\mathbf{X}_b$, $\mathbf{K}^c_{NN}$, $\mathbf{K}^c_{M_cM_c}$, $\mathbf{K}^c_{M_cN}$, $\mathbf{m}^c$, $\mathbf{S}^c$}
        
        \State $\text{ELL}^{\text{reg}} \leftarrow$\Call{ExpectedLogLikelihoodRegression}{$\mathbf{Y}^{r}$, $\boldsymbol{\mu}_{\mathbf{f}^r} , \boldsymbol{\Sigma}_{\mathbf{f}^r}$, $\boldsymbol{\Sigma}_d^2$, $\mathbf{X}$}
        
        \State $\text{ELL}^{\text{cls}} \leftarrow$\Call{ExpectedLogLikelihoodClassification}{$\mathbf{Y}^{c}$, $\boldsymbol{\mu}_{\mathbf{f}^c} , \boldsymbol{\Sigma}_{\mathbf{f}^c}, \mathbf{X}$}

        \State $\text{KL}_U \leftarrow \text{KL}(q_{\lambda}(\mathbf{U}^{c}) \parallel p(\mathbf{U}^{c} \mid \mathbf{Z}^{c})) + \text{KL}(q_{\gamma}(\mathbf{U}^{r}) \parallel p(\mathbf{U}^{r} \mid \mathbf{Z}^{r}))$
        \State $\text{ELBO} \leftarrow \text{ELL}^{\text{reg}} + \text{ELL}^{\text{cls}} - \text{KL}_U - \text{KL}(q_{\phi}(\mathbf{X}) \parallel p(\mathbf{X}))$
        \State Compute gradient of $-ELBO$ w.r.t $\{\theta, \psi, \phi, \lambda, \gamma, \mathbf{Z}\}$
        \State Update $\{\theta, \psi, \phi, \lambda, \gamma, \mathbf{Z}\}$ using gradient descent with learning rate $lr $
        \State Append $-ELBO$to$ELBO$list
    \EndFor
    \State \textbf{return}$\{\theta^*, \psi^*, \phi^*, \lambda^*, \gamma^*, \mathbf{Z}^*\}$
\EndProcedure

\end{algorithmic}
\label{alg:train}
\end{algorithm}

\subsubsection{Prediction}
After optimizing the hyperparameters using a training dataset, we aim to employ LDGD to predict the labels for a new set of measurements, denoted as ${\mathbf{Y}^r}^*$. This set of measurements constitutes the test set. We assume the absence of corresponding labels ${\mathbf{Y}^c}^*$ for these new data points. In essence, we employ the test data points to infer their latent features and subsequently attempt to decode these latent features to determine the distribution over the corresponding labels.

The variational posterior for the latent variables $\mathbf{X}$ was introduced in Equation~\eqref{eq:x_variational_aprox}. Given a new test point ${\mathbf{Y}^r}^*$, we seek to compute the posterior distribution $p(\mathbf{X}^* \mid {\mathbf{Y}^r}^*, \mathbf{Y}^r, \mathbf{Y}^c)$, which represents our belief about the latent variables after observing the new test data. This can be approximated as $q(\mathbf{X}^*) = \prod_{i=1}^{N_{test}} \mathcal{N}(\mathbf{x}_i^* \mid \boldsymbol{\mu}_*, \mathbf{s}_* \mathbb{I}_Q)$.

To find the optimum $\boldsymbol{\mu}_*$ and $\mathbf{s}_*$, we aim to minimize the negative of the lower bound concerning these new variational parameters. During this process, the inducing points and inducing variables are held constant. Therefore, the objective function for optimizing the variational parameters associated with a test point is formulated as follows:
\begin{equation*}
    \arg\max_{\boldsymbol{\mu}_*, \mathbf{s}_*} \left\{ \sum_{d} \text{ELL}_{*,d}^{\text{reg}} - \text{KL}\left(q_\phi(\mathbf{X}^*) \parallel p(\mathbf{X}^*)\right) \right\}.
    \label{eq:ELBO_test}
\end{equation*}
Using this result, we can effectively decode and determine $p({\mathbf{Y}^c}^* \mid \mathbf{X}^*)$.

\subsubsection{Fast LDGD}
In LDGD, we need to run an iterative optimization in the prediction or test phase, making it less suitable in a real-time scenario. In this section, we introduce Fast LDGD, in which we employ a neural network to estimate the variational parameters for latent space directly. To this end, we modify the variational distributions introduced in Equation~\eqref{eq:variational_distributions} and estimate the posterior $p(\mathbf{X} \mid \mathbf{Y}^{r}, \mathbf{Y}^{c})$ with
\begin{equation}
    q_{\phi}(\mathbf{X}) = \prod_{i=1}^{N} \mathcal{N}(
        \mathbf{x}_i \mid 
        \mu(\mathbf{y}_i^{r}), 
        s(\mathbf{y}_i^{r}) \mathbb{I}_Q
        ), 
        \label{eq:x_variational_approx}
\end{equation}

Where $\mu(\mathbf{y}_i^{r})$ and $s(\mathbf{y}_i^{r})$ are neural network outputs with trainable parameters $\mathbf{\phi}$.  The architecture of the neural network can be chosen based on the data. In this approach, the number of the model parameters depends on the continuous data dimensionality and does not depend on the number of data points. In contrast, in LDGD, the number of variational parameters in Equation~\eqref{eq:variational_distributions} grows linearly with the number of training samples. Thus, this approach is beneficial when dealing with a decoding problem (classification) with a relatively larger number of data points. Our goal in the training phase is to minimize the negative of the ELBO by optimizing neural network parameters, inducing points, inducing variables' variational parameters, kernel parameters, and model parameters. By achieving this, we ensure the neural network can estimate a suitable approximation to the true posterior given the continuous data. Once trained, the model can instantly provide estimates of the posterior's mean and covariance for a test point without additional optimization. This capability facilitates real-time inference.

This approach has several advantages. First, by directly mapping from observations to latent variables' variational parameters, we bypass the need for iterative optimization to determine $\mathbf{\mu}_i$ and $\mathbf{s}_i$ (introduced in Equation~\eqref{eq:variational_distributions}) for each test data point. Consequently, this significantly reduces computational time, enabling real-time predictions, which is crucial in time-sensitive applications. 

\section{Experiments and Results}
In this section, we study different attributes of LDGD by applying it to various datasets. In particular, we examine its capability to draw low-dimensional data representations, decode accurately (classification), and generate data points. We investigate these attributes using both synthetic and real-world datasets. By utilizing synthetic data, we can conduct a comprehensive investigation of the model attributes under controlled conditions. The real-world datasets include the Oil Flow ~\citep{bishop1993analysis}, Iris~\citep{misc_iris_53}, and MNIST datasets \citep{deng2012mnist}. Through these datasets, we further explore how the model's attributes can extend to handling complex datasets, thereby underscoring the unique benefits of LDGD. These benefits include its ability to infer meaningful features from high-dimensional data, address variability present in the data, and achieve optimal low-dimensional data representation. 

We also compare LDGD modeling results with different variants of GPLVM. Through these comparisons, we aim to underscore LDGD's distinctive advantages. In applying this model to classification tasks, we evaluate it across three datasets using classification metrics to demonstrate the model's robustness in class label prediction. Additionally, we compared LDGD classification accuracy with two other supervised GPLVM variants, SLLGPLVM \citep{jiang2012supervised} and SGPLVM \citep{gao2010supervised}, to illustrate its superiority in classification. We explore the model's data generation capability using the MNIST dataset, showing that LDGD can effectively replicate the complex patterns of handwritten digits. We also compare LDGD and fast LDGD with variational auto-encoders, where we highlight LDGD and fast LDGD's key features over variational auto-encoders, including optimal discovery of low dimension representation, robustness with limited dataset and inference process. These analyses will illustrate LDGD's advantages in analyzing complex and high-dimensional datasets.

\subsection{Datasets}
\subsubsection{Synthetic data}
For the synthetic data, we utilize a ''moon-like'' dataset, which is a two-dimensional dataset resembling a pair of crescent moons depicted in Figure~\ref{fig: synthetic}. We transfer this two-dimensional data into much higher dimensional space. To do this, we use two different transformations. The first transformation involves a linear transformation, where the original data is multiplied by a randomly generated matrix of size $D \times 2$, where $D$ is set to 5, 15, or 20. In the second transformation, we double the dimension of the data by adding white noise data dimensions. For instance, if the data dimension is 5, we generate five channels of independent white noise with unit variance and append it to the data. The added dimensions do not carry class-specific information, making inference and classification tasks hard. For each specified dimensionality, we generated 500 synthetic data samples, divided them into two classes with 250 samples per class, and labeled them synthetic-10, synthetic-30, and synthetic-40.

 \subsubsection{Oil Flow Dataset}
 \label{subsec:synthetic_data}
The Oil Flow dataset is a high-dimensional dataset consisting of 12-dimensional feature vectors. These data points are recorded to capture various factors representing oil flow dynamics within a pipeline \citep{bishop1993analysis}. The dataset embodies three distinct phases of flow: horizontally stratified, nested annular, and homogeneous mixture flow. These three phases represent the conditions under which oil, gas, and water can coexist and move through a pipeline. Despite the high dimensionality of the dataset, the crucial information about the flow conditions can be effectively described by two main variables: the fraction of water and the fraction of oil in the flow. This property of the dataset makes it an extensively used benchmark for evaluating the effectiveness of dimensionality reduction techniques, showcasing their ability to simplify and interpret complex, high-dimensional data in fluid dynamics and pipeline management. A total of 1,000 samples are included in the dataset. 

\subsubsection{Iris Dataset}
The Iris dataset is widely used in machine learning, covering morphological variations across three categories of iris flowers, namely Setosa, Versicolor, and Virginica \citep{misc_iris_53}. It comprises 150 data points distributed equally among the species, each detailing four critical features reflecting the flowers' physical characteristics, specifically petal and sepal sizes. The dataset's clear class distinctions and simple structure make it an ideal example of how our dimensionality reduction technique finds meaningful low-dimensional representation.

\subsubsection{MNIST Dataset}
The MNIST (Modified National Institute of Standards and Technology) dataset is popularly used in machine learning and image processing \citep{deng2012mnist}. The dataset comprises 70,000 images of handwritten digits (0 through 9), and it serves as a fundamental benchmark for evaluating the performance of various machine-learning algorithms. Each image is represented in a gray-scale format with a resolution of 28x28 pixels, equating to 784 dimensions when flattened into a vector form. 

\subsection{Performance Assessment}
\subsubsection{Synthetic Data}
\begin{figure}[t]
    \centering
    \includegraphics[width=\linewidth]{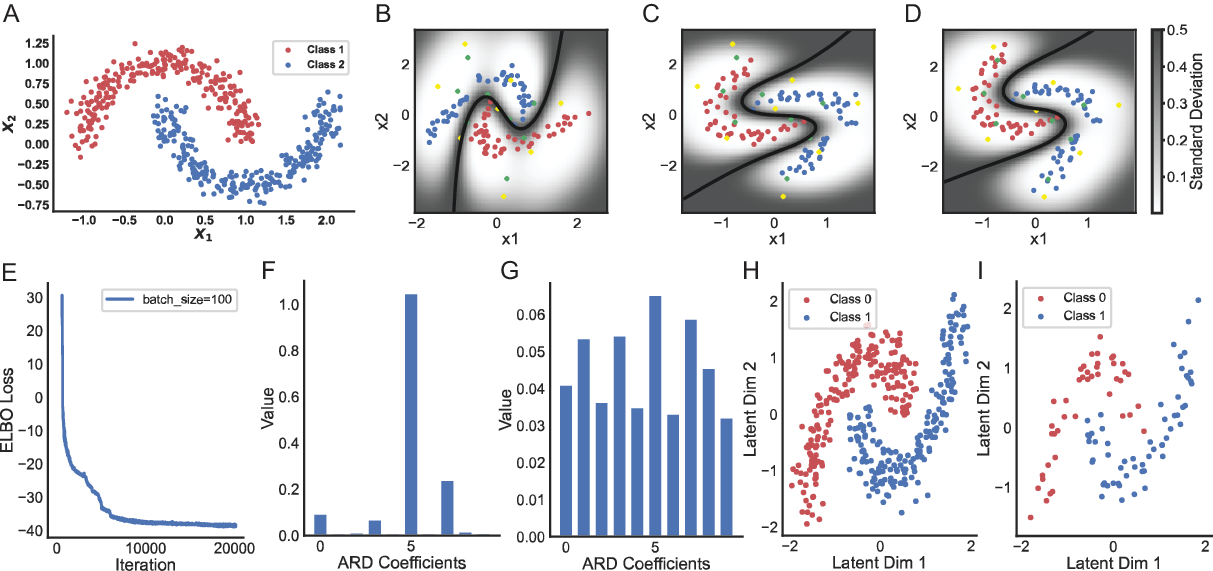}
    \caption{Visualization of Dimensionality Reduction on Synthetic Data. (A) Displays the initial two dimensions of the moon-like dataset. (B-D) Illustrates the latent space heatmap representation across different synthetic data dimensions: 10 dimensions (B), 20 dimensions (C), and 40 dimensions (D). Red and blue points represent class 1 samples ($y^c=0$) and class 2 samples ($y^c=1$), respectively. Green crosses indicate classification-inducing points, while yellow crosses denote regression-inducing points, with five inducing points used for each. The green data points are more uniformly distributed over the space as it is constructing the $\mathbf{Y}^r$. The yellow ones are aligned on two sides of the decision boundary as they are more needed for a correct classification. The heatmap visualizes the model's uncertainty level (posterior variance). (E) Shows the training curve (ELBO loss) for synthetic data with ten dimensions, where the latent space is also set to 10 dimensions. (F-G) Depicts the ARD coefficients for the classification kernel (F) and the regression kernel (G). The trained coefficients highlight that the model selects two dimensions to represent a 10-dimensional space in a lower-dimensional setting for decoding labels and employs almost all dimensions to reconstruct data in the original space. (H) displays a scatter plot of the training points in the lower-dimensional space for the two most dominant dimensions, while (I) shows the scatter plot for test points where the labels are unknown.}
    \label{fig: synthetic}
\end{figure}

\paragraph{Lower Dimension Representation.} We argue that two key advantages of our model are its ability to infer meaningful information in high-dimensional data and, more importantly, quantify the validity of its probabilistic inference. To demonstrate this, LDGD is applied on the synthetic-10, synthetic-30, and synthetic-40 datasets, described in Section~\ref{subsec:synthetic_data}. We run LDGD with two different settings. In the first setting, we assumed $Q$, the latent dimension, is 2. This choice of dimension is helpful to properly visualize the inferred $\mathbf{X}$. Figure~\ref{fig: synthetic}-B, \ref{fig: synthetic}-C, and \ref{fig: synthetic}-D respectively show inferred $\mathbf{X}$ for the synthetic-10, synthetic-30, and synthetic-40 dataset. The inferred points (posterior mean of $\mathbf{X}$) represent a rotated version of the initial data shown in Figure~\ref{fig: synthetic}-A. Here, our pipeline converges to a similar data representation independent of the observed data dimension. The variance of the predictive distribution \( p(\mathbf{Y}^c \mid X) \) serves as a measure of uncertainty in label prediction. The heatmap, employing a binary colormap, visualizes the uncertainty levels across different regions. The inferred posterior variance reflects higher confidence in areas with denser data concentrations (segments) and lower confidence in regions where these concentrations approach one another.

We use the second setting to highlight LDGD's capability to infer the latent dimension representing observed data and corresponding labels optimally. This can be done by adjusting ARD coefficients in the training step. For this setting, we use the synthetic-10 dataset and set LDGD's latent dimensions to 10, equal to the data dimension, with 25 inducing points for each classification and regression path.
 The model was trained using 80\% of the dataset, selected randomly. The training process aimed to minimize Equation~\eqref{eq:elbo_all}, as described in Section~\ref{section:ELBO}. Figure~\ref{fig: synthetic}-E shows values of ELBO loss over training iterations. For the classification, the ARD coefficients indicate LDGD converges to 2-D representation aligned with what is embedded in the data (Figure~\ref{fig: synthetic}-F). For the regression, the ARD coefficients suggest that all dimensions are employed for data generation (Figure~\ref{fig: synthetic}-G). The ARD coefficients in each of the two paths, along with the inferred state, demonstrate how the framework tries to reconstruct both paths through the same latent structure. The inferred $\mathbf{X}$ shows what the model requires for classification and what is needed for reconstructing data. The regression ARD coefficients display larger values for the dimensions identified by the classification ARD coefficients, suggesting that these dimensions play a significant role in data reconstruction. However, additional dimensions are also necessary. This is anticipated, as the expanded data in each dimension is a weighted sum of the original data, necessitating the involvement of other latent dimensions as well.

 Subsequently, with the optimized hyperparameters, we optimized the partial ELBO introduced in Equation~\eqref{eq:ELBO_test} to find the estimated posterior over latent variables for the remaining 20\% of the data without any knowledge of the labels. Figure~\ref{fig: synthetic}-I demonstrates that, even for test data with unknown labels, the model could discern the intrinsic pattern hidden within the high dimensions even for test data with unknown labels.
 
\paragraph{Classification Results.}
To assess the model's performance in classification tasks, LDGD was evaluated using 5-fold cross-validation. In each fold, we optimized the parameters using 80\% of the synthetic dataset's data points, with the remaining 20\% serving as the test set, where labels were assumed to be unknown. Twenty-five inducing points within a 10-dimensional latent manifold were utilized for LDGD. Subsequently, the model's precision, recall, and F-measure were evaluated on the test set. These results are detailed in Table~\ref{tab:classification_results1}. The model demonstrated accurate label prediction on the test set, showcasing another critical feature of LDGD: its capability to decode labels.

\begin{table}[t]
\centering
\begin{tabular}{lccc}
\hline
\textbf{Dataset} & \textbf{Precision} & \textbf{Recall} & \textbf{F1 Score} \\
\hline
Synthetic-10& 1.0& 1.0& 1.0 \\
Synthetic-20& 1.0& 1.0& 1.0 \\
Synthetic-40& 1.0& 1.0& 1.0 \\
\hline
\end{tabular}
\caption{Classification performance of LDGD for Synthetic data}
\label{tab:classification_results1}
\end{table}

\subsubsection{Oil Flow, Iris, and MNIST Dataset}
\begin{figure}[t]
    \centering
    \includegraphics[width=\linewidth]{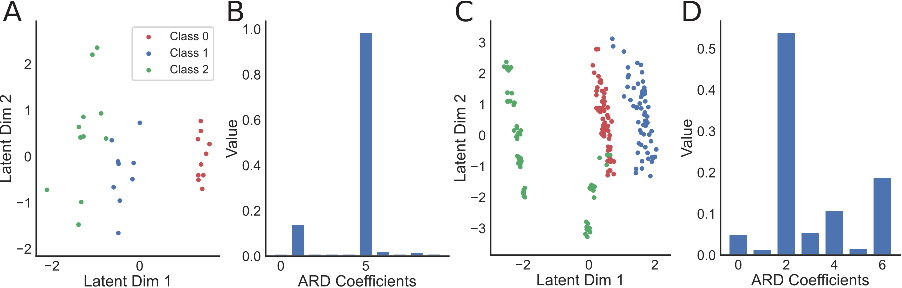}
    \caption{Comparative analysis of latent space representation in Iris and Oil Flow dataset. (A) reveals the most dominant latent dimension for the Iris dataset, identified through ARD coefficients. (B) illustrates the scatter plot of the two dominant dimensions in the latent space for the Iris dataset, highlighting the data's intrinsic clustering. The vertical and horizontal error bars show the variance at each data point in latent space. (C) displays the most dominant latent dimension for the Oil dataset, as determined by ARD coefficients. (D) presents the scatter plot of the two dominant dimensions in the latent space for the Oil Flow dataset, showcasing its unique distribution.}
    \label{fig:toy_datasets}
\end{figure}

\paragraph{Lower Dimension Representation.}
To investigate the LDGD application further in inferring low-dimensional representation of data, we applied it to the Oil Flow, Iris, and MNIST datasets. We used an ARD kernel with seven dimensions for the Oil Flow and Iris datasets. The model exhibits low sensitivity to the choice of latent dimension size; therefore, we have selected 7, which exceeds the dimensionality of the Iris dataset but remains below that of the Oil Flow dataset. We also assume we have ten inducing points for the classification path and the same number for the regression path. We optimized the parameters using 80\% of the data along with their labels and inferred the latent space for the remaining 20\% without label information. The results indicate that the trained ARD coefficients suggest an optimal use of about one dimension for the Iris dataset and two dimensions for the Oil Flow dataset (see Figure~\ref{fig:toy_datasets}). We can see there are different levels of certainty in each axis; for instance, in Figure~\ref{fig:toy_datasets}-B, we can see much higher confidence in x-axis, corresponding to $5^{th}$ element in ARD space, indicating its significant role in the model's predictions. On the other hand, the second largest ARD coefficient is less significant, leading to lower confidence in y-axis. For Iris, the dimension of $\mathbf{X}$ is larger than the observed data. Though this is not the case the framework is designed for, we observed that ARD only keeps two coefficients and pushes other coefficients toward zero. Though this does not reflect any attributes of the LDGD, it shows the robustness of the pipeline and its principled representation of the latent data.

Given the MNIST dataset's larger feature space, we used 20 dimensions for the latent space and 150 inducing points for each path. The model utilized almost all the available dimensions for this dataset, highlighting the necessity of high dimensions to classify and reconstruct such a large dataset. LDGD's representation in 2-D may not exhibit as much distinct separation as T-SNE due to its generative nature and probabilistic embeddings that prioritize capturing the global data structure over local neighborhood relationships.

\begin{figure}[t]
\centering
\includegraphics[width=\linewidth]{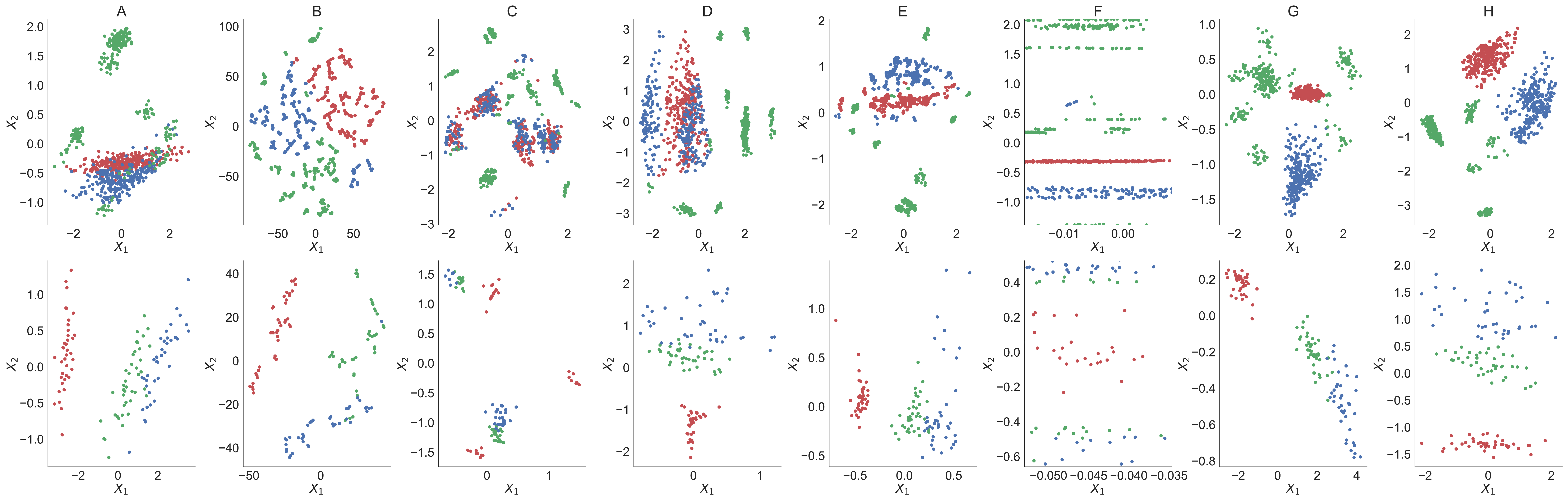}
\caption{Two-dimensional visualization of dataset projections using various dimensionality reduction techniques. The top row displays projections of the Oil Flow dataset, while the bottom row shows the Iris dataset. Techniques used include (A) PCA, (B) t-SNE, (C) GPLVM, (D) Bayesian GPLVM, (E) FGPLVM, (F) SGPLVM, (G) SLLGPLVM, and (H) LDGD. We observed SLLGPLVM and LDGD inference in low dimension reflecting the class label, ending to separate regions for different data classes. With LDGD, we can even have a one-dimensional representation of data. Thus, we might even get a better reduction rate using LDGD.}
\label{fig:latent_vis_iris_oil}
\end{figure}

In parallel, we applied PCA, t-SNE, GPLVM, Bayesian GPLVM, FGPLVM, SGPLVM, and SLLGPLVM to the Oil Flow and Iris datasets. The latent variable's dimensions were fixed to two across all models for fair comparisons. SGPLVM, SLLGPLVM, and LDGD utilize labels during the training process, while others are unsupervised and agnostic to the labels. Figure~\ref{fig:latent_vis_iris_oil} illustrates the latent representations generated by LDGD compared to those obtained using other models. The Iris dataset, being less complex, allowed almost all models to find a lower-dimensional representation that demonstrates a clear distinction between species. So, LDGD's performance is similar to other models. For the Oil Flow dataset, LDGD and SLLGPLVM effectively separate the three flow phases, while some other methods exhibit partial overlap between different categories. Unlike SLLGPLVM, LDGD incorporates inducing points within its pipeline, which allows LDGD to scale better than SLLGPLVM. Consequently, a direct comparison for the MNIST dataset is not feasible since SLLPLVM is ill-equipped to handle larger datasets.

\paragraph{Classification Results.}
Here, we aim to demonstrate the model's ability to decode or classify real-world datasets. We have applied it to the Iris and Oil Flow with seven latent dimensions and ten inducing points. Then, we applied LDGD with 20 latent variables and 150 inducing points for classification and regression for the MNIST dataset. The evaluation criteria included classification accuracy, precision, recall, and F1 score. The results are provided in Table~\ref{tab:classification_results_real_world}. As we see, the model performance is as good as the state of the arts in the iris and oil dataset, and the performance in the MNIST dataset is not far from the other image classifier methods like convolutional neural network (CNN) methods but comparable to them \cite{alvear2019improving}. The classification results in datasets with different input dimensions and different data points further demonstrate the model potential in the feature extraction and classification step. 

\begin{table}[t]
\centering
\begin{tabular}{lcccc}
\hline
\textbf{Dataset} & \textbf{Accuracy} & \textbf{Precision} & \textbf{Recall} & \textbf{F1 Score} \\
\hline
Iris & 1.00 & 1.00 & 1.00 & 1.00 \\
Oil Flow & 0.99 & 1.00 & 0.99 & 0.99 \\
MNIST & 0.95 & 0.94 & 0.93 & 0.94 \\
\hline
\end{tabular}
\caption{Classification performance of LDGD for Iris, Oil Flow, and MNIST dataset}
\label{tab:classification_results_real_world}
\end{table}

To quantitatively compare the classification capabilities of our model, we benchmarked its performance against several state-of-the-art models. While our model and others perform well on the simpler Iris dataset, LDGD stands out for its unique ability to handle the complexity of the MNIST dataset. Consequently, our comparison focuses on the Oil Flow dataset to provide a fair comparison of LDGD's performance with its competitors.

Table~\ref{tab:classification_results} summarizes the classification results, where LDGD consistently outperforms other models across all measures. This confirms LDGD's effectiveness in extracting latent features and accurately predicting labels. The model's superior classification performance results from its adaptive feature extraction capabilities. These empirical findings highlight LDGD's dual role as both an efficient tool for dimensionality reduction and a powerful discriminative model. The model's achievement of high classification metrics across various datasets underscores its adaptability and potential for use in numerous practical machine-learning scenarios, even with larger datasets. Our results for MNIST, a comparatively large dataset for Gaussian process-based models, demonstrate LDGD's scalability. The LDGD model surpasses both SLLGPLVM and SGPLVM in terms of scalability. The scalability is attributed mainly to integrating batch training, which enables efficient handling of large datasets without compromising the model's performance or accuracy.

\begin{table}[t]
\centering
\begin{tabular}{lcccc}
\hline
\textbf{Model} & \textbf{Accuracy} & \textbf{Precision} & \textbf{Recall} & \textbf{F1 Score} \\
\hline
LDGD & 0.99 & 1.00 & 0.99 & 0.99 \\
SGPLVM & 0.93 & 0.92 & 0.91 & 0.92 \\
SLLGPLVM & 0.95 & 0.94 & 0.93 & 0.94 \\
\hline
\end{tabular}
\caption{Classification performance of LDGD for the Oil Flow dataset compared to other models}
\label{tab:classification_results}
\end{table}

\begin{figure}[t]
\centering
\includegraphics[width=\linewidth]{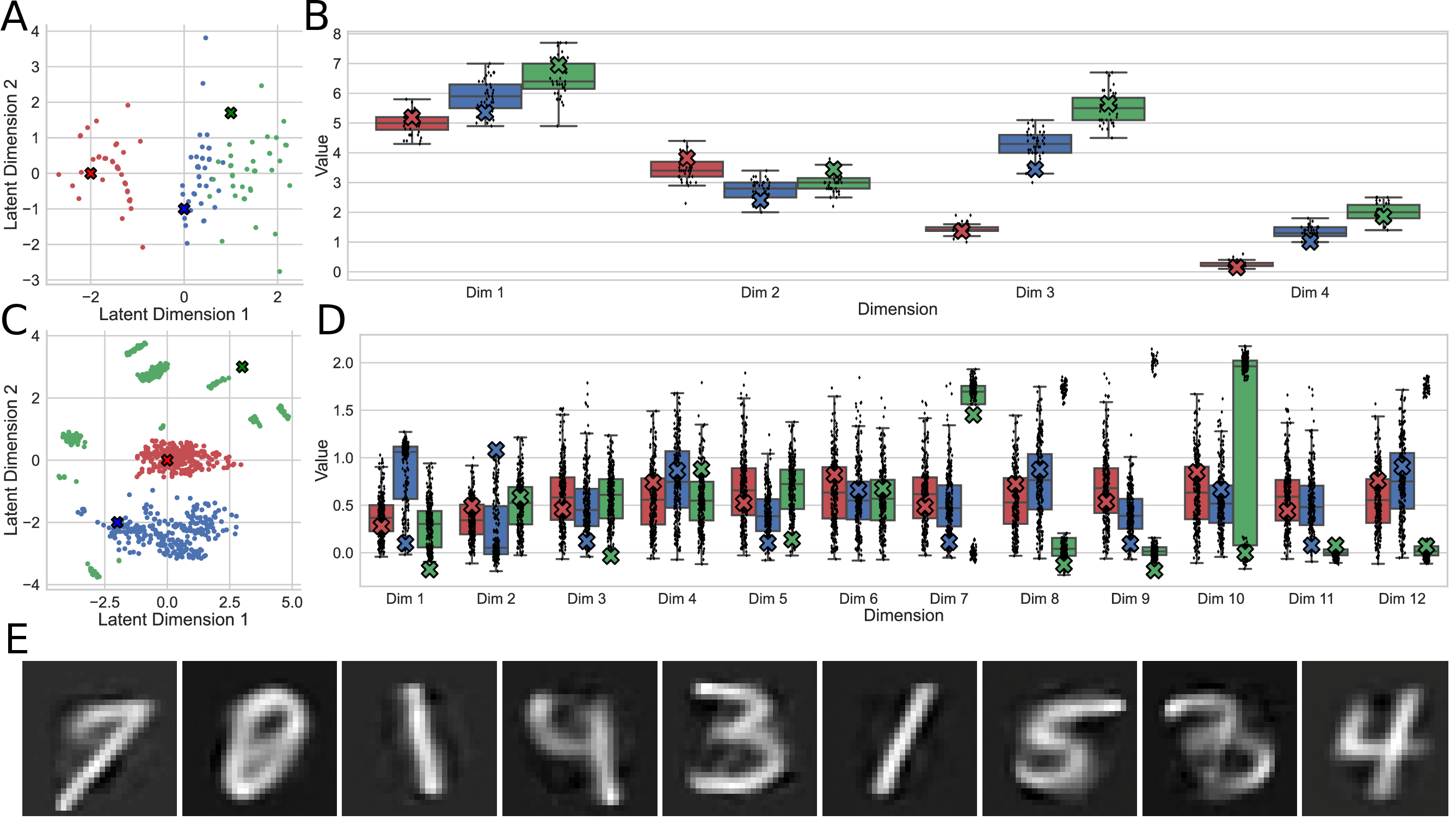}
\caption{LDGD Data Generation Analysis. (A-D) These figures present scatter plots of data points in the 2D latent space for the Iris (A) and the Oil Flow dataset (C). For each dataset, a sample point near each cluster is randomly selected (marked with a cross) to illustrate the model's generative capabilities. The corresponding high-dimensional reconstructions of these starred points are shown for the Iris dataset (B) and the Oil Flow dataset (D), alongside the real values of other points. (E) displays nine randomly chosen test points in the latent space for the MNIST dataset and their reconstructed images through the model's generative path. The generated images properly reconstruct corresponding digits.}
\label{fig:datageneration}
\end{figure}

\paragraph{LDGD Data Generation.}
So far, our focus has been on inferring the latent space and then utilizing LDGD for label prediction. It is noteworthy that LDGD is capable of generating continuous samples given a label. Figure~\ref{fig:datageneration} displays three sample data points, each selected randomly near a class in the latent space within the Iris and Oil Flow datasets. As observed, the reconstructed samples in high-dimensional space lie close to the actual data points. This demonstrates how LDGD training effectively balances the label and continuous paths, ensuring that the generated data accurately represents its respective classes. Additionally, Figure~\ref{fig:datageneration}-C illustrates nine sample digits generated randomly from the MNIST dataset, highlighting the model's ability to produce samples from more complex data.

\subsection{LDGD, Fast LDGD, and Variational Auto-Encoders}
\begin{figure}[t]
\centering
\includegraphics[width=\linewidth]{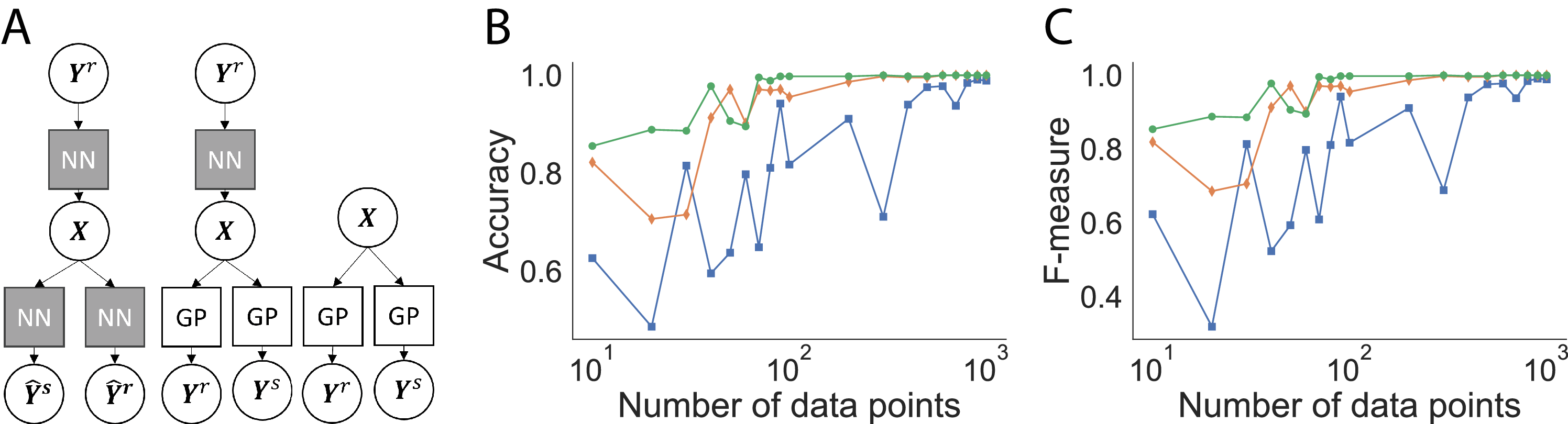}
\caption{Comparison of LDGD, Fast LDGD, and VAE Models. (A) shows the graphical models for VAE, fast LDGD, and LDGD from left to right. (B, C) illustrate the relationship between the number of training samples (x-axis) and the performance metrics on the test set, specifically accuracy (B) and F-measure (C), where the blue line represents VAE performance, red indicates LDGD, and green denotes fast LDGD performance.}
\label{fig:Compare_vae}
\end{figure}

While LDGD might resemble a variational auto-encoder (VAE), fundamental differences between our proposed framework and VAE models must be highlighted further. In VAEs, the latent manifold has Gaussian priors, similar to LDGD and fast LDGD, where the posterior over latent variables are approximated using a Gaussian distribution. However, the distinctions lie in the relationship between the latent manifold, labels, and high-dimensional data. LDGD employs a probabilistic framework, whereas VAEs utilize a deterministic framework (neural networks). Fast LDGD occupies a middle ground, inferring $\mathbf{X}$ through a neural network while leveraging a Gaussian Process (GP) model for label prediction. Another key difference resides in the loss functions used: VAEs incorporate mean squared error (MSE), categorical entropy, and Kullback-Leibler (KL) divergence, whereas LDGD relies on the ELBO. Figure~\ref{fig:Compare_vae}.A shows the probabilistic graphs for these three methods. As we can see in LDGD, we are trying to find the posterior over \(\mathbf{X}\) given \(\mathbf{Y}^r\) and \(\mathbf{Y}^c\).

LDGD has multiple advantages over the VAE model, making it a more suitable model for analyzing high-dimensional data, specifically when the size of dataset is limited. Using LDGD, we can identify the optimal dimension of the latent process as a part of learning. In contrast, we would have to tune the number of dimensions as a hyperparameter using the VAE model. Inferencing and training in LDGD is not a data-greedy process; we can build the model with much smaller data than needed for VAE model training. Compared to a point estimate in VAE, Bayesian inferencing helps us have higher-order statistics, such as confidence in our prediction of a label or data point, giving us a better sense of our predictions.

Another key advantage is the inherent robustness of LDGD against overfitting, a common pitfall in many machine learning models. LDGD's use of Bayesian inferencing inherently guards against overfitting, making LDGD particularly suited for dealing with small datasets. This framework doesn't merely fit the model to the data; it enables the model to correctly infer and adapt to the data, even when the dataset is small. To assess the model's performance in such scenarios, we have used the synthetic-20 dataset. We used a different amount of training data with the same amount of test data and compared VAE, LDGD, and fast LDGD. As shown in Figure~\ref{fig:Compare_vae}-B and \ref{fig:Compare_vae}-C, in small datasets, LDGD and fast LDGD's performance in classification is significantly better. For large datasets, the performance is comparable. These advantages collectively position LDGD as a mere alternative and a significant advancement over traditional encoder-decoder models, especially in high-dimensional data analysis.

\section{Discussion}
In this research, we developed different modeling steps of LDGD, including its training, inferencing, and decoding~(classification). In developing the framework, we introduced new approaches, including a shared latent variable with doubly stochastic variational inference with trainable inducing points. To assess different attributes of the model, we applied the framework in the Synthetic, Iris, and Oil Flow datasets.

The modeling results in both synthetic and real-world data suggest LDGD's capability to correctly infer latent space and identify the latent space's proper dimension. The classification results using LDGD are on par with the state-of-the-art models, and in some of the data points, it even surpasses other approaches. 

We pointed to different attributes and potential advantages of LDGD. We argued its capabilities in identifying the optimal dimension of the latent process. For instance, in synthetic data, we showed that the framework would suggest 2-D dimensional latent space, which matches the intrinsic dimensions of the original data. This can be done using an ARD kernel where its coefficients are trained during the model training. This is a remarkable feature of the model, which is not present in other algorithms such as VAEs. We also pointed to the LDGD's expressiveness and ability to classify data labels precisely. We showed that LDGD can be a perfect Iris and Oil Flow dataset classifier, surpassing state-of-the-art approaches. We can see LDGD as a pipeline with adaptive feature extraction and a classifier with features capable of reconstructing the data. Our results demonstrated its capability to process larger datasets, such as MNIST, efficiently, which was achieved by adopting batch training, the inducing point concept, and a doubly variational strategy. Specifically, this approach highlighted LDGD's scalability to larger datasets, a feature distinctly lacking in models like SGPLVM and SLLGPLVM. We also explored the generative prowess of LDGD, that is, its ability to create data from high-dimensional spaces. Our experiments confirmed that LDGD  can adeptly generate samples from the Iris, Oil Flow, and MNIST datasets. The generated samples from MNIST were remarkably coherent.

We observed a couple of challenges with LDGD as well. A core challenge is hyperparameter selection, such as determining the number of inducing points. As with many in machine learning, the model's performance hinges on hyperparameter choices. In Gaussian Process-based models, the number of inducing points is critical; too few may lead to underfitting and inadequate data complexity representation, while too many can escalate computational demands and risk overfitting. Finding the optimal number often requires a dataset-specific balance, posing practical challenges. Although LDGD improves scalability, it may struggle with extremely large or high-dimensional datasets due to the inherent complexity of Bayesian models and Gaussian Processes, limiting its feasibility in some real-time applications.

LDGD's reliance on label quality for dimensionality reduction means its performance is contingent on label accuracy. The model's efficiency may diminish in cases of scarce, inaccurate, or biased labels. Despite its proficiency in uncertainty quantification, the interpretability of complex models like LDGD can be daunting, especially for users without a machine learning background. The contribution of various modeling steps to create the outcome might make the pipeline hard to digest. Furthermore, while LDGD shows promise for particular datasets, its performance across diverse data types, such as time series, images, or text, has not been thoroughly tested, leaving its applicability in these areas yet to be fully established.

Several improvements are suggested to overcome these limitations and extend the applicability of LDGD. These include the development of methods for the automatic selection of inducing points, such as Bayesian optimization and cross-validation approaches, as well as extensions of the model to handle time series data. Several modifications are suggested to enhance the LDGD model for time series analysis. Firstly, incorporate temporal dynamics by integrating autoregressive components, state space models, or using time series-specific kernels in Gaussian Processes \citep{wong2000mixture, fard2023latent, gu2023mamba}. Secondly, implementing methods to handle sequential data, possibly using recurrent neural network structures like LSTMs \citep{yu2019review} or GRUs \citep{dey2017gate} within the LDGD framework. Thirdly, explore time-variant inducing points that adapt to changes in time series data, improving the model's ability to capture temporal dynamics. Parallel computing, GPU acceleration, and sparse variational frameworks are also recommended to enhance algorithmic efficiency and scalability.

In general, it is essential to establish standard benchmarks and evaluation methodologies to advance the field of dimensionality reduction and ensure the efficacy of techniques like the LDGD model. Additionally, it's crucial to continuously benchmark the enhanced model against the latest state-of-the-art models in dimensionality reduction. This ongoing comparison will ensure the model's competitiveness and help identify potential areas for further improvement.

Our future work will focus on extending LDGD for specific applications, such as datasets we record during neuroscience experiments, which are generally small in size and show complex variability \citep{ziaei2023bayesian}. We also aim to enhance algorithmic efficiency and conduct robustness and generalizability studies. In summary, the LDGD model offers promising advancements in dimensionality reduction and classification, with wide-ranging applications and potential for significant contributions to various fields. However, realizing its full potential requires continuous research and development to address its current limitations and expand its applicability in addressing complex real-world problems.

\section{Conclusion}
In this research, we showcased the utilities of non-parametric models for dimensionality reduction, feature discovery, and decoding. Despite the emergence of powerful machine-learning tools, we still lack principled tools capable of dealing with intrinsic stochasticity and variability in data or scenarios where the size of available data is limited. The LDGD framework addresses these needs and achieves comparable or even higher prediction accuracy and generative capability. With LDGD, we can study data from domains such as neuroscience, which are generally complex and limited in size. The ideas in LDGD can be extended to more realistic data, such as time series, thus requiring further model development, which is the core of our future research. 

\section{Code Availability}
The source code for our model and usage examples is freely available to the research community for further exploration and development at our GitHub repository: \url{https://github.com/Navid-Ziaei/LDGD}. This initiative promotes transparency, reproducibility, and collaborative advancements within the field.

\newpage
\section{Appendixes}
\appendix
\section{Exploring the Connection between PPCA and GPLVM }
\label{appendix_ppca}
The GPLVM is developed based on the probabilistic principal component analysis (PPCA) framework. In PPCA, the observed data vector \( \mathbf{y}_i \) for each observation \( i \) is modeled as a linear transformation of the corresponding latent variable \( \mathbf{x}_i \) with additive Gaussian noise \( \boldsymbol{\epsilon}_i \):
\begin{equation*}
    \mathbf{y}_i = \mathbf{W}\mathbf{x}_i + \boldsymbol{\epsilon}_i,
\end{equation*}
where \( \mathbf{y}_i \in \mathbb{R}^D \) denotes the observations, \( \mathbf{x}_i \in \mathbb{R}^Q \) represents the low-dimensional representation, and \( \mathbf{W} \in \mathbb{R}^{D \times Q} \) is the transformation matrix mapping from the low-dimensional space to the high-dimensional space. The prior distribution over $\mathbf{X}$ is assumed to be Gaussian with a zero mean and an identity covariance matrix:
\begin{equation*}
    p(\mathbf{X}) = \prod_{i=1}^{N} \mathcal{N}(\mathbf{x}_i \mid \mathbf{0}, \mathbb{I}_Q).
\end{equation*}
Consequently, the likelihood of $\mathbf{Y}$ given the transformation matrix $\mathbf{W}$ is:
\begin{equation*}
    p(\mathbf{Y} \mid \mathbf{W}) = \prod_{i=1}^{N} \mathcal{N}(\mathbf{y}_i \mid \mathbf{0}, \mathbf{C}),
\end{equation*}
with the covariance matrix $\mathbf{C}$ defined as:
\begin{equation*}
    \mathbf{C} = \mathbf{W}\mathbf{W}^T + \sigma^2 \mathbb{I}_D.
\end{equation*}
In the dual formulation of PPCA, we consider the model from the perspective of the observed data:
\begin{equation*}
    \mathbf{y}_{:,d} = \mathbf{X}\mathbf{w}_{:,d} + \boldsymbol{\epsilon}_d,
\end{equation*}
where \( \mathbf{X} \in \mathbb{R}^{N\times Q} \) represents the low-dimensional representation of all samples, \( \mathbf{y}_{:,d} \) is the \(d\)th feature in high-dimensional space for all samples, and \( \mathbf{w}_{:,d} \) is the \(d\)th column of \( \mathbf{W} \), which is the linear transformation matrix. The prior distribution over \( \mathbf{W} \) is assumed to be Gaussian with a zero mean and an identity covariance matrix:
\begin{equation*}
    p(\mathbf{W}) = \prod_{d=1}^{D} \mathcal{N}(\mathbf{w}_{:,d} \mid \mathbf{0}, \mathbb{I}_Q).
\end{equation*}
Here, the likelihood of $\mathbf{Y}$ given the latent variables $\mathbf{X}$ is:
\begin{equation*}
    p(\mathbf{Y} \mid \mathbf{X}) = \prod_{d=1}^{D} \mathcal{N}(\mathbf{y}_{:,d} \mid \mathbf{0}, \mathbf{K}),
\end{equation*}
where $\mathbf{K}$ is the covariance matrix reflecting a linear relationship to $\mathbf{X}$:
\begin{equation*}
    \mathbf{K} = \mathbf{X}\mathbf{X}^T + \sigma^2 \mathbb{I}_N.
\end{equation*}
To accommodate the nonlinear relationships present in many real-world datasets, we can extend the dual PPCA formulation using kernel functions $k_{\theta}(\mathbf{x}, \mathbf{x}')$. This formulation enables us to express the covariance matrices through kernel functions as follows:
\begin{equation*}
    \mathbf{K} = \mathbf{K}_{NN} + \sigma^2 \mathbb{I}_N,
\end{equation*}
where \(\mathbf{K}_{NN}\) denotes the covariance matrix, with each element \(k_{ij}\) in the \(i\)th row and \(j\)th column calculated by the kernel function \(k_{\theta}(\mathbf{x}_i, \mathbf{x}_j)\), reflecting the covariance between two points \(\mathbf{x}_i\) and \(\mathbf{x}_j\) in the input space. The incorporation of kernel functions into the PPCA model gives rise to the Gaussian Process Latent Variable Model (GPLVM), allowing for the capture of nonlinear relationships within the data. This enhancement transforms the model into a more adaptable framework, capable of accommodating the data's inherent complexity, as described by:
\begin{equation*}
    p(\mathbf{Y} \mid \mathbf{X}) = \prod_{d=1}^{D} \mathcal{N}(\mathbf{y}_{:,d} \mid \mathbf{0}, \mathbf{K}_{NN} + \sigma^2 \mathbb{I}_N),
\end{equation*}
where \(\mathbf{Y}\) represents the high-dimensional data, \(\mathbf{X}\) denotes the latent low-dimensional representation, and \(\mathbf{K}_{NN}\) is the covariance matrix derived from the kernel function, capturing the nonlinearities in the data. Thus, GPLVM extends the linear dimensionality reduction approach of PPCA to a nonlinear setting, offering a more powerful tool for uncovering the underlying structure in complex datasets.

\section{Evidence Lower Bound for GPLVM}
\label{apendix_LDGD_elbo}
In this appendix, we demonstrate that maximizing the ELBO (Equation~\ref{eq:elbo_all}) is equivalent to minimizing the KL divergence between the true posterior distribution and the variational posterior distribution. Equations~\eqref{eq:posterior_with_induce} and \eqref{eq:variational_posterior} depict LDGD's true and approximated posteriors, respectively. Let's denote the true posterior as $\hat{P} \triangleq p(\mathbf{F}^r, \mathbf{F}^c, \mathbf{U}^{r}, \mathbf{U}^c, \mathbf{X} \mid
\mathbf{Y}^r, \mathbf{Y}^c)$ and approximated posterior as \(\hat{Q} \triangleq q(\mathbf{F}^{r}, \mathbf{F}^{c}, \mathbf{U}^{r}, \mathbf{U}^{c}, \mathbf{X}) = q(\mathbf{F}^c, \mathbf{U}^{c}) q(\mathbf{F}^r, \mathbf{U}^{r}) q_{\phi}\). We begin by computing the KL divergence between the exact posterior and our approximation:

\begin{align}
\label{eq:apx_kl_posterior1}
    \textit{KL}&(\hat{Q} \parallel \hat{P}) 
    = \int{\hat{Q} 
        \log \frac{
        \hat{Q}}
        {\hat{P}} 
        d\mathbf{F}^{r}\, d\mathbf{F}^{c}\, d\mathbf{U}^{r}\, d\mathbf{U}^{c}\, d\mathbf{X}} 
\end{align}

The posterior distribution of LDGD, given the observed data \( \mathbf{Y}^r \) and \( \mathbf{Y}^c \), the latent functions \( \mathbf{F}^r \) and \( \mathbf{F}^c \), the inducing variables \( \mathbf{U}^{r} \) and \( \mathbf{U}^{c} \), and the latent features \( \mathbf{X} \), is defined as:
\begin{align*}
    \hat{P} =& \frac{
    p(\mathbf{F}^{r}, \mathbf{F}^{c}, \mathbf{U}^{r}, \mathbf{U}^{c}, \mathbf{X}, \mathbf{Y}^{r}, \mathbf{Y}^{c})}
    {p(\mathbf{Y}^r, \mathbf{Y}^c)}.
\end{align*}
We then substitute this into Equation \eqref{eq:apx_kl_posterior1}:
\begin{align}
    \textit{KL}(\hat{Q} \parallel \hat{P})
    =& \int{
         \hat{Q} 
        \log \frac{
        p(\mathbf{Y}^r, \mathbf{Y}^c)\hat{Q}}
        {p(\mathbf{F}^r, \mathbf{F}^c, \mathbf{U}^{r}, \mathbf{U}^c, \mathbf{X},
        \mathbf{Y}^r, \mathbf{Y}^c)} 
        d\mathbf{F}^{r}\, d\mathbf{F}^{c}\, d\mathbf{U}^{r}\, d\mathbf{U}^{c}\, d\mathbf{X}} \notag \\
    =& \log p(\mathbf{Y}^r, \mathbf{Y}^c) + \int{
         \hat{Q} 
         \log \frac{
        \hat{Q}}
        {p(\mathbf{F}^r, \mathbf{F}^c, \mathbf{U}^{r}, \mathbf{U}^c, \mathbf{X},
        \mathbf{Y}^r, \mathbf{Y}^c)} 
        d\mathbf{F}^{r}\, d\mathbf{F}^{c}\, d\mathbf{U}^{r}\, d\mathbf{U}^{c}\, d\mathbf{X}} \notag \\
        \triangleq& \log p(\mathbf{Y}^r, \mathbf{Y}^c) + \text{A}
\end{align}
We focus on the integral term. We utilize Equations \eqref{eq:ldgd_joint_dist} and \eqref{eq:variational_posterior} to factorize the joint distribution and the approximated posterior, respectively:
\begin{align*}
    \text{A}=& \int{
         \hat{Q} 
         \log \frac{
        \hat{Q}}
        {p(\mathbf{F}^r, \mathbf{F}^c, \mathbf{U}^{r}, \mathbf{U}^c, \mathbf{X},
        \mathbf{Y}^r, \mathbf{Y}^c)} 
        d\mathbf{F}^{r}\, d\mathbf{F}^{c}\, d\mathbf{U}^{r}\, d\mathbf{U}^{c}\, d\mathbf{X}} \notag \\
    =&\int{
        \hat{Q} 
         \log \frac{
            q_{\phi}(\mathbf{X}) 
            q_{\lambda}(\mathbf{U}^{c})
            q_{\gamma}(\mathbf{U}^{r})}
        {p(\mathbf{Y}^{c} \mid \mathbf{F}^{c})
        p(\mathbf{U}^{c})
        p(\mathbf{Y}^{r} \mid \mathbf{F}^{r})
        p(\mathbf{U}^{r})
        p(\mathbf{X})} 
        d\mathbf{F}^{r}\, d\mathbf{F}^{c}\, d\mathbf{U}^{r}\, d\mathbf{U}^{c}\, d\mathbf{X}}\notag \\
    =& E_{\hat{Q}}\left[
         \log \frac{
            q_{\phi}(\mathbf{X}) 
            q_{\lambda}(\mathbf{U}^{c})
            q_{\gamma}(\mathbf{U}^{r})}
        {p(\mathbf{Y}^{c} \mid \mathbf{F}^{c})
        p(\mathbf{U}^{c})
        p(\mathbf{Y}^{r} \mid \mathbf{F}^{r})
        p(\mathbf{U}^{r})
        p(\mathbf{X})} 
        \right]
\end{align*}
Given that the KL divergence is always non-negative, we have:
\begin{align*}
    \log p(\mathbf{Y}^r, \mathbf{Y}^c) \geq 
        E_{\hat{Q}} \left[\log \frac
        {p(\mathbf{Y}^{c} \mid \mathbf{F}^{c})
        p(\mathbf{U}^{c})
        p(\mathbf{Y}^{r} \mid \mathbf{F}^{r})
        p(\mathbf{U}^{r})
        p(\mathbf{X})}{
            q_{\phi}(\mathbf{X}) 
            q_{\lambda}(\mathbf{U}^{c})
            q_{\gamma}(\mathbf{U}^{r})}
        \right] \triangleq \textit{ELBO}
\end{align*}
Thus, we have identified a lower bound for the marginal likelihood, known as the Evidence Lower Bound (ELBO). An alternative approach to achieve the same result involves marginalizing the joint distribution, followed by multiplying and dividing the joint distribution by the joint posterior approximation:
\begin{align*}
    p(\mathbf{Y}^{r}, \mathbf{Y}^{c}) 
    =&\int
        p(\mathbf{Y}^{r}, \mathbf{Y}^{c}, \mathbf{F}^{r}, \mathbf{F}^{c}, \mathbf{U}^{r}, \mathbf{U}^{c}, \mathbf{X}) \, 
        d\mathbf{F}^{r} \, d\mathbf{F}^{c} \, d\mathbf{U}^{r} \, d\mathbf{U}^{c} \, d\mathbf{X} \notag \\
    =&\int
        \frac{\hat{Q}}{\hat{Q}}p(\mathbf{Y}^{r}, \mathbf{Y}^{c}, \mathbf{F}^{r}, \mathbf{F}^{c}, \mathbf{U}^{r}, \mathbf{U}^{c}, \mathbf{X}) \, 
        d\mathbf{F}^{r} \, d\mathbf{F}^{c} \, d\mathbf{U}^{r} \, d\mathbf{U}^{c} \, d\mathbf{X} \notag \\
    &\int 
        \hat{Q}
         \frac{
            p(\mathbf{Y}^{c} \mid \mathbf{F}^{c})
            p(\mathbf{U}^{c})
            p(\mathbf{Y}^{r} \mid \mathbf{F}^{r})
            p(\mathbf{U}^{r})
            p(\mathbf{X})
        }{
            q_{\phi}(\mathbf{X}) 
            q_{\lambda}(\mathbf{U}^{c}) 
            q_{\gamma}(\mathbf{U}^{r})
        }
        \, d\mathbf{F}^{r} \, d\mathbf{F}^{c} \, d\mathbf{U}^{r} \, d\mathbf{U}^{c} \, d\mathbf{X} \notag \\
    =& E_{\hat{Q}}\left[
        \frac{
            p(\mathbf{Y}^{c} \mid \mathbf{F}^{c})
            p(\mathbf{U}^{c})
            p(\mathbf{Y}^{r} \mid \mathbf{F}^{r})
            p(\mathbf{U}^{r})
            p(\mathbf{X})
        }{
            q_{\phi}(\mathbf{X}) 
            q_{\lambda}(\mathbf{U}^{c}) 
            q_{\gamma}(\mathbf{U}^{r})
        }
    \right]
\end{align*}
Using Jensen inequality we can find the evidence lower-bound (ELBO):
\begin{align*}
    \log p(\mathbf{Y}^{r}, \mathbf{Y}^{c}) 
    \geq&
    E_{q(\mathbf{F}^{r}, \mathbf{F}^{c}, \mathbf{U}^{r}, \mathbf{U}^{c}, \mathbf{X})}\left[
        \log \frac{
            p(\mathbf{Y}^{c} \mid \mathbf{F}^{c})p(\mathbf{U}^{c})
            p(\mathbf{Y}^{r} \mid \mathbf{F}^{r})p(\mathbf{U}^{r})
            p(\mathbf{X})
        }{
            q_{\phi}(\mathbf{X})
            q_{\lambda}(\mathbf{U}^{c})
            q_{\gamma}(\mathbf{U}^{r})
        }
    \right] \triangleq \textit{ELBO}
\end{align*}
Having calculated the ELBO through two approaches, we now proceed to decompose the ELBO into simpler terms:
\begin{align*}
    \textit{ELBO} =& 
    E_{q(\mathbf{F}^{c}, \mathbf{U}^{c}, \mathbf{X})}\left[
        \log \frac{
            p(\mathbf{Y}^{c} \mid \mathbf{F}^{c})p(\mathbf{U}^{c})
        }{
            q_{\lambda}(\mathbf{U}^{c})
        }
    \right] + 
    E_{q(\mathbf{F}^{r}, \mathbf{U}^{r}, \mathbf{X})}\left[
        \log \frac{
            p(\mathbf{Y}^{r} \mid \mathbf{F}^{r})p(\mathbf{U}^{r})
        }{
            q_{\lambda}(\mathbf{U}^{c})
        }
    \right] \notag \\
    &- \textit{KL}(q_{\phi}(\mathbf{X}) \parallel p(\mathbf{X})) \notag \\
    = 
    & E_{q_{\phi}(\mathbf{X})} 
    \left[
        E_{p(\mathbf{F}^{c} \mid \mathbf{U}^{c}, \mathbf{X}) 
        q_{\lambda}(\mathbf{U}^{c})}
        \left[\log p(\mathbf{Y}^{c} \mid \mathbf{F}^{c})\right] 
    \right]+ \notag \\
    & E_{q_{\phi}(\mathbf{X})} 
    \left[
        E_{p(\mathbf{F}^{r} \mid \mathbf{U}^{r}, \mathbf{X}) 
        q_{\gamma}(\mathbf{U}^{r})}
        \left[\log p(\mathbf{Y}^{r} \mid \mathbf{F}^{r})\right]
    \right] - \notag \\
    & \textit{KL}(q_{\lambda}(\mathbf{U}^{c}) \parallel p(\mathbf{U}^{c}))
    - \textit{KL}(q_{\lambda}(\mathbf{U}^{r}) \parallel p(\mathbf{U}^{r}))
    - \textit{KL}(q_{\phi}(\mathbf{X}) \parallel p(\mathbf{X})) \notag \\
    =& \text{ELL}^{\text{reg}} + \text{ELL}^{\text{cls}} - \text{KL}_{u}^{c}- \text{KL}_{u}^{r}- \text{KL}_{X}.
\end{align*}

\section{Computation of Predictive Distribution}
\label{appendix_predictive_dist}
In this appendix, we detail the calculation of the predictive distribution. Given that the calculation for both continuous and labeled data follows the same process, we focus on the continuous case for simplicity. The LDGD framework employs a Gaussian prior for the inducing variables, formalized as $p(\mathbf{u}_{:,d}^{r} \mid \mathbf{Z}) = \mathcal{N}(\mathbf{u}_{:,d}^{r} \mid \mathbf{0}, \mathbf{K}_{M_rM_r}^{r})$. Moreover, the variational distribution within this model is specified as $q_\gamma(\mathbf{U}^r) = \prod_{d=1}^{D} \mathcal{N}(\mathbf{u}_{:,d}^{r} \mid \mathbf{m}_d^r, \mathbf{S}_{d}^{r})$. Utilizing the transformation matrix $A$, defined as $A \triangleq {\mathbf{K}_{NM_r}^r} {\mathbf{K}_{M_rM_r}^{r}}^{-1}$, the distribution of function values, conditional on the inducing points and latent variables, is then expressed as:
\begin{equation*}
p(\mathbf{f}_{:,d}^{r} \mid \mathbf{u}_{:,d}^{r}, \mathbf{X}) = \mathcal{N}(\mathbf{A} \mathbf{u}_{:,d}^{r}, \mathbf{K}_{NN}^r - \mathbf{A} \mathbf{K}_{M_rM_r}^{r} \mathbf{A}^T).
\end{equation*}
This formulation leverages Gaussian distribution properties for the marginal distribution.  Subsequently, the predictive distribution, represented as $q_\gamma(\mathbf{f}_{:,d}^{r} \mid \mathbf{X})$, is derived by integrating out the inducing variables. Given that both terms within the integral are Gaussian distributions, this integration has an analytical solution, allowing for straightforward computation:
\begin{align*}
q_\gamma(\mathbf{f}_{:,d}^{r} \mid \mathbf{X}) 
&\triangleq \int 
    p(\mathbf{f}_{:,d}^{r} \mid \mathbf{u}_{:,d}^{r}, \mathbf{X}) 
    q_\gamma(\mathbf{u}_{:,d}^{r}) \, 
    d\mathbf{u}_{:,d}^{r} = \mathcal{N}(
        \mathbf{f}_{:,d}^{r} \mid 
        \boldsymbol{\mu}_f^r, \boldsymbol{\Sigma}_f^r), \\
    \boldsymbol{\mu}_f^r&=\mathbf{A}^T\mathbf{m}_d^r, \\
    \boldsymbol{\Sigma}_f^r&=\mathbf{K}_{NN}^{r} 
    + \mathbf{A}^T (\mathbf{S}^{r}_{d} - {\mathbf{K}_{M_rM_r}^{r}}) \mathbf{A}.
\end{align*}

For stabilizing the learning of variational parameters, we utilize the whitening trick proposed by \cite{murray2010slice}. Let $\mathbf{L}^r$ denote the Cholesky factor of the covariance matrix of the prior over inducing variables, such that ${\mathbf{K}_{M_rM_r}^{r}} = \mathbf{L}^r(\mathbf{L}^r)^T$. We introduce whitened parameters $\mathbf{m}^r_d = \mathbf{L}^r\hat{\mathbf{m}}^r_d$ and $\mathbf{W}^r_d = \mathbf{L}^r\hat{\mathbf{W}}^r_d$, where $\hat{\mathbf{m}}^r_d$ and $\hat{\mathbf{W}}^r_d$ are our free parameters. Here, $\mathbf{W}^r_d$ and $\hat{\mathbf{W}}^r_d$ are upper triangular matrices. The covariance matrix of the variational distribution is computed as $\mathbf{S}^r_d = \mathbf{W}^r_d(\mathbf{W}^r_d)^T$. Given these definitions and transformations, the mean and covariance of the predictive distribution are expressed as 
$\mathbf{\mu}^r_f = \hat{\mathbf{A}}^T \hat{\mathbf{m}}^r_d$ 
and 
$\mathbf{\Sigma}^r_f = \mathbf{K}_{NN}^{r} + \hat{\mathbf{A}}^T\left(\hat{\mathbf{W}}^{r}_{d} (\hat{\mathbf{W}}^{r}_{d})^T - \mathbb{I} \right)\hat{\mathbf{A}}$
, where 
$\hat{\mathbf{A}}={L^r}^{-1}\mathbf{K}^r_{M_rN}$ . The introduction of whitened parameterization and the transformation of the mean vector simplify computation, ensuring numerical stability and computational efficiency. This approach is applicable to categorical data (labels) as well:
\begin{align*}
q_\lambda(\mathbf{f}_{:,k}^{c} \mid \mathbf{X}) 
&\triangleq \int 
    p(\mathbf{f}_{:,k}^{c} \mid \mathbf{u}_{:,k}^{c}, \mathbf{X}) 
    q_\lambda(\mathbf{u}_{:,k}^{c}) \, 
    d\mathbf{u}_{:,k}^{c} = 
    \mathcal{N}(\mathbf{f}_{:,k}^{c} \mid \boldsymbol{\mu}_f, \boldsymbol{\Sigma}_f), \\
    \boldsymbol{\mu}^c_f &= B^T \mathbf{m}_{k}^{c}, \\
    \boldsymbol{\Sigma}^c_f&=\mathbf{K}_{NN}^{c} 
    + B^T 
    (\mathbf{S}^{c}_{k} - {\mathbf{K}_{M_cM_c}^{c}}),
    B
\end{align*}
where $B \triangleq {\mathbf{K}_{M_cM_c}^{c}}^{-1} \mathbf{K}_{M_cN}^{c}$. Using whitened parameters, we can express it as $\boldsymbol{\mu}_f^c = \hat{\mathbf{B}}^T \hat{\mathbf{m}}_k^c$ and  $\boldsymbol{\Sigma}_f^c = \mathbf{K}_{NN}^{c} + \hat{\mathbf{B}}^T (\hat{\mathbf{W}}_k^c(\hat{\mathbf{W}}_k^c)^T - \mathbb{I})\hat{\mathbf{B}}$, where $\hat{\mathbf{B}} = (\mathbf{L}^c)^{-1}\mathbf{K}_{M_cN}^{c}$ and $\mathbf{K}_{M_cM_c}^{c} = \mathbf{L}^c(\mathbf{L}^c)^T$.

Implementation details for the predictive distribution are provided in Algorithm~\ref{alg:predictive_dist}. The calculation process is identical for both continuous and labeled data; hence, we omit the superscripts 'r' and 'c' used in the main text to differentiate between the two data types.
\begin{algorithm}[ht]
\caption{Calculate Predictive Distribution}
\begin{algorithmic}[1]
\Require 
\Statex Sampled latent variables $\mathbf{X}_b$
\Statex Kernel matrices $\mathbf{K}_{NN}, {\mathbf{K}_{MM}}, \mathbf{K}_{MN}$
\Statex Whitened variational mean $\hat{\mathbf{m}}$
\Statex Whitened variational covariance  $\hat{\mathbf{S}}=\hat{\mathbf{W}}\hat{\mathbf{W}}^T$ ($\hat{\mathbf{W}}$ is upper triangular)

\Function{PredictiveDistribution}{$\mathbf{X}_b$, $\mathbf{K}_{NN}$, ${\mathbf{K}_{MM}}$, $\mathbf{K}_{MN}$, $\hat{\mathbf{m}}$, $\hat{\mathbf{S}}$}
    \State Compute kernel matrices $\mathbf{K}_{NN}, {\mathbf{K}_{MM}}, \mathbf{K}_{MN}$ using kernel function
    \State Add jitters to ${\mathbf{K}_{MM}}$
    \State Decompose ${\mathbf{K}_{MM}}$ into $LL^T$ using Cholesky decomposition    
    \State Compute interpolation term $\hat{\mathbf{A}} \leftarrow L^{-1} \mathbf{K}_{MN}$
    \State Calculate predictive mean $\mathbf{\mu}_f \leftarrow \hat{\mathbf{A}}^T \hat{\mathbf{m}}$
    \State Calculate predictive covariance $\mathbf{\Sigma}_f \leftarrow \hat{\mathbf{A}}^T (\hat{\mathbf{W}}\hat{\mathbf{W}}^T - I)\hat{\mathbf{A}}$

    \State $\mathbf{\Sigma}_f \leftarrow \mathbf{K}_{NN}^{r} + \mathbf{\Sigma}_f$ 
    
    \State \textbf{return} $\mathbf{\mu}_f$ and $\mathbf{\Sigma}_f$
\EndFunction

\end{algorithmic}
\label{alg:predictive_dist}
\end{algorithm}

\section{Expected Log-Likelihood Implementation}
\label{appendix_ell}
The implementation details for calculating \( \text{ELL}^{\text{reg}} \) based on Equation \eqref{eq:ell_reg_estimate} and \( \text{ELL}^{\text{cls}} \) based on Equation \eqref{eq:ell_cls_estimate} are provided in Algorithms~\ref{alg:ell_reg} and \ref{alg:ell_cls}, respectively.

\begin{algorithm}[h]
\caption{Calculate Expected Log-Likelihood (ELL) for Regression}
\begin{algorithmic}[1]
\Require 
\Statex Observations $\mathbf{y}$
\Statex Predictive distribution $p(y_{i,d}^{r} \mid f_{d}^{r} (\mathbf(X)))= \mathcal{N}(f_{d}^{r} (\mathbf(X)), \mathbf{\sigma}_d^2)$
\Statex Noise variance $\sigma_d^2$
\Statex Sampled points $\mathbf{X}^{(j)}$, $j=1,\ldots,J$
\Function{ExpectedLogLikelihoodRegression}{$\mathbf{y}$, $\mathbf{\mu}_{\mathbf{f}_{:,d}^{r}}$, $\mathbf{\Sigma}_{\mathbf{f}_{:,d}^{r}}$, $\mathbf{\sigma}_d^2$, $\mathbf{X}^{(j)}$}
    \State Initialize $ELL_{i,d}^{\text{reg}} \leftarrow 0$
    \For{$j = 1$ to $J$}
        \State Compute $\mathbf{\mu}_{\mathbf{f}_{:,d}^{r}}(x^{(j)}_{i})$ and $\mathbf{\Sigma}_{\mathbf{f}_{:,d}^{r}}(x^{(j)}_{i})$
        \State Calculate $A(x^{(j)}_{i}) \leftarrow -0.5 \left[ \log 2\pi + \log \mathbf{\sigma}_d^2 +\frac{(y_{i,d}^{r}-\mathbf{\mu}_{\mathbf{f}_{:,d}^{r}}(x^{(j)}_{i}))^2+\mathbf{\Sigma}_{\mathbf{f}_{:,d}^{r}}(x^{(j)}_{i})}{\mathbf{\sigma}_{d}^2} \right]$
        \State Update $ELL_{i,d}^{\text{reg}} \leftarrow ELL_{i,d}^{\text{reg}} + A(x^{(j)}_{i})$
    \EndFor
    \State Normalize $ELL_{i,d}^{\text{reg}} \leftarrow -\frac{1}{2J} \times ELL_{i,d}^{\text{reg}}$

    \State \textbf{return} $ELL_{i,d}^{\text{reg}}$
\EndFunction

\end{algorithmic}
\label{alg:ell_reg}
\end{algorithm}

\begin{algorithm}[H]
\caption{Calculate Expected Log Probability using Gauss-Hermite Quadrature}
\begin{algorithmic}[1]
\Require 
\Statex Observations $\mathbf{y}$
\Statex Predictive distribution $q_{\gamma}(\mathbf{f}_{k}^{c} \mid \mathbf{X})= \mathcal{N}\left(\mathbf{f}_{k}^{c} \mid \mathbf{\mu}_{f_{k}^{c}}(\mathbf{X}), \mathbf{\Sigma}_{f_{k}^{c}}(\mathbf{X}) \right)$
\Statex Gauss-Hermite Quadrature nodes $\mathit{Loc}$
\Statex Gauss-Hermite Quadrature weights $\omega$

\Function{ExpectedLogLikelihoodClassification}{$\mathbf{y}$, $f(\mathbf{x})$, $\mathit{Loc}$, $\mathbf{\omega}$}
    \State $\mathbf{y} \leftarrow 2\mathbf{y} - 1$  \Comment{Normalize $\mathbf{y}$ to $\{-1, 1\}$}
    
    \State $\mathit{Loc}_{shifted} \leftarrow \sqrt{2 \mathbf{\Sigma}_{f_{k}^{c}}(\mathbf{X})} \times \mathit{Loc} + \mathbf{\mu}_{f_{k}^{c}}(\mathbf{X})$
    \Comment{Shift and scale quadrature locations}
    
    \State $log\_probs \leftarrow \log \Phi(\mathit{Loc}_{shifted} \cdot \mathbf{y})$
    \Comment{Compute log probabilities using normal CDF $\Phi$}
    
    \State $\mathbb{E}_{q_{\gamma}(\mathbf{f}_{k}^{c} \mid \mathbf{X})}[\log P(\mathbf{y} \mid f_{k}^{c}] \approx \frac{1}{J\sqrt{\pi}} \sum_{j=1}^{J} log\_probs^{(j)} \times \omega^{(j)}$
    \Comment{Approximate expected log probability}
    
    \State \textbf{return} Approximated expected log probability
\EndFunction

\end{algorithmic}
\label{alg:ell_cls}
\end{algorithm}

\newpage
\vskip 0.2in
\bibliography{main}

\end{document}